\RequirePackage{fix-cm}
\documentclass[twocolumn]{svjour3}          
\smartqed  

\usepackage{graphicx}
\usepackage{amsmath,amssymb} 
\usepackage{algorithm}
\usepackage{algorithmic}
\usepackage{color}

\def\vect#1{\mbox{\boldmath $#1$}}
\def\x{{\vect{x}}}

\def\1{{\mathbf 1}}

\def\n{{\mathbf n}}

\def\X{{\mathbf X}}

\def\betab{{\boldsymbol\beta}}

\def\alphab{{\boldsymbol\alpha}}

\def\D{{\mathbf D}}
\def\DD{{\mathcal D}}
\def\s{{\mathbf s}}
\def\b{{\mathbf b}}

\def\n{{\mathbf n}}

\def\B{{\mathbf B}}

\def\W{{\mathbf W}}
\def\d{{\mathbf d}}
\def\E{{\mathbb E}}

\def\s{{\mathbf s}}
\def\k{{\mathbf k}}

\def\Drond{{\mathcal D}}

\def\S{{\mathbf S}}

\def\d{{\mathbf d}}

\def\b{{\mathbf b}}
\def\u{{\mathbf u}}

\def\Real{{\mathbb R}}

\def\U{{\mathbf U}}
\def\u{{\mathbf u}}

\def\argmin{\operatornamewithlimits{arg\,min}}

\def\defin{\stackrel{\vartriangle}{=}}
\newcommand{\normd}[1]{\| #1 \|_2 ^2}
\newcommand{\normu}[1]{\| #1 \|_1}
\newcommand{\normF}[1]{\| #1 \|_F ^2}

\journalname{International Journal of Computer Vision}

\begin{document}
\title{Dictionary Learning for Deblurring and Digital Zoom}

\author{Florent Couzinie-Devy \and  Julien Mairal \and Francis Bach \and Jean Ponce
}

\institute{F. Couzinie-Devy \at
	    Willow Project-Team, Laboratoire d'Informatique de l'\'Ecole Normale Sup\'erieure (ENS/INRIA/CNRS UMR 8548), \\
	    23, avenue d'Italie, CS 81321, 75214 Paris Cedex 13, France.\\
            \email{couzinie@ens.fr}           
           \and
	    J. Mairal \at
	   Departement of Statistics, University of California, Berkeley \\
	   \#301, Evans Hall, CA 94720-3860. \\
	   \email{julien@stat.berkeley.edu}
    	  \and
	    F. Bach \at
	  Sierra Project-Team, Laboratoire d'Informatique de l'\'Ecole Normale Sup\'erieure (ENS/INRIA/CNRS UMR 8548), \\
	    23, avenue d'Italie, CS 81321, 75214 Paris Cedex 13, France.\\
	  \email{francis.bach@inria.fr}    
            \and
	   J. Ponce \at
	    Willow Project-Team, Laboratoire d'Informatique de l'\'Ecole Normale Sup\'erieure (ENS/INRIA/CNRS UMR 8548), \\
	    23, avenue d'Italie, CS 81321, 75214 Paris Cedex 13, France. \\
	  \email{jean.ponce@ens.fr}
}
\date{05/10/2011}
\maketitle
\begin{abstract}
This paper proposes a novel approach to image deblurring and digital zooming using
sparse local models of image appearance.  These models, where small image
patches are represented as linear combinations of a few elements drawn from
some large set (dictionary) of candidates, have proven well
adapted to several image restoration tasks.  A key to their success has been to
learn dictionaries adapted to the reconstruction of small image patches. In contrast,
 recent works have proposed instead to learn dictionaries which are not only adapted 
to data reconstruction, but also tuned for a specific task. We introduce here such an approach to
deblurring and digital zoom, using pairs of blurry/sharp (or
low-/high-resolution) images for training, as well as an effective stochastic
gradient algorithm for solving the corresponding optimization task. Although
this learning problem is not convex, once the dictionaries have been learned,
the sharp/high-resolution image can be recovered via convex optimization at
test time. Experiments with synthetic and real data demonstrate the
effectiveness of the proposed approach, leading to state-of-the-art performance
for non-blind image deblurring and digital zoom.

\end{abstract}
\section{Introduction}
With recent advances in sensor design, the quality of the signal
output by digital reflex and hybrid/bridge cameras is remarkably
high. Point-and-shoot cameras, however, remain susceptible to noise at
high sensitivity settings and/or low-light conditions, and this
problem is exacerbated for mobile phone cameras with their small
lenses and sensor areas. Photographs taken with a long exposure time are less noisy but
may be blurry due to
movements in the scene or camera shake. Likewise, although the image
resolution of modern cameras keeps on increasing, there is a clear
demand for high-quality digital zooming from amateur and professional
photographers, whether they crop their family vacation pictures or use
footage from camera phones in newscasts. Thus, the classical image restoration
problems of denoising, deblurring, multi-frame super-resolution and digital zooming
(also called single-image super-resolution) are still of acute and in fact growing importance,
and they have received renewed attention lately with the
emergence of computational photography (e.g.,
\cite{fergus,glasdner,levin}).

The image deblurring problem is naturally ill-posed: Indeed, perfect low-pass
filters remove all high-frequen\-cy information from images.  They are
non-invertible operators, and different sharp images can give rise to the same
blurry one.  Thus, an appropriate image model is required to
regularize the deblurring process. Several explicit priors for natural images
have been proposed in the past for different tasks in image restoration. Early work relied on various smoothness
assumptions, or image decompositions on
fixed bases such as wavelets~\cite{mallat}. More recent approaches
include non-local means filtering~\cite{buades}, learned sparse models~\cite{elad,YiMa,mairalNonLocal}, 
piecewise linear estimator~\cite{yu2010},
Gaussian scale mixtures~\cite{portilla}, fields of experts \cite{roth}, 
kernel regression~\cite{takeda}, and
block matching with 3D filtering (BM3D)~\cite{dabov}. Pairs of low-/high-resolution images have also been used as an {\em implicit} image prior in
digital zooming tasks~\cite{freeman}, and combining the
exemplar-based approach with image self-similarities at different scales has
recently led to impressive results~\cite{glasdner}.

We propose in this paper to build on several of these ideas with a new
approach to non-blind image deblurring (the blur kernel is assumed to
be fixed and known) and digital zooming. Like Freeman et al.~\cite{freeman}, we use training pairs of blurry/sharp or
low-/high-resolution image patches readily available for these tasks
to learn our model parameters. We also exploit learned sparse local
models of image appearance, as in~\cite{elad,YiMa}, which have been known
to be very effective for several image reconstruction tasks.
   Our method shares some ideas with the work of Yang et al.~\cite{YiMa}, but 
 our formulation combines several novelties that improves the results:

-~Whereas the approach of \cite{YiMa} is purely generative (this model learns how to
  simultaneously reconstruct pairs of low- and high-resolution patches), our
approach learns how to reconstruct a high-resolution patch \emph{given} a
low-resolution one. In essence, the difference is the same as between
generative and discriminative models in machine learning.

-~We present a novel formulation for non-blind image
  deblurring and digital zooming, combining a \emph{linear predictor} with \emph{dictionary learning},
 and show with extensive experiments on both synthetic and real data that our approach 
is competitive with the state of the art for these two tasks.

-~We adapt the stochastic gradient descent of~\cite{mairalPAMI}
  for solving the corresponding learning problem allowing the use of large databases of training patches (typically
several millions).

\textbf{Notation.} We define for $p\geq 1$ the $\ell_p$ norm of a vector $\x$ in $\Real^m$ as $\|\x\|_p=(\sum_{i=1}^m|\x[i]|^p)^{1/p}$, where $\x[i]$ denotes the
$i$-th coordinate of $\x$. We denote the Frobenius norm of a matrix $\X$ in $\Real^{m\times n}$ by $\|\X\|_F=(\sum_{i=1}^m\sum_{j=1}^n|\X[i,j]|^2)^{1/2}$.

\section{Related Work} \label{sec:related}

\subsection{Deblurring and Digital Zoom}
Blur is a common image degradation, and the literature on the
subject is quite large (see, e.g., \cite{GEM,fergus,BOA,foi2006,levin,takeda}).
Most existing methods assume a shift-invariant blur operator 
such that a blurry image $\B$ can be modelled as the
convolution of the sharp image $\S$ with a fixed blur kernel $\k$: 
\begin{equation}
  \B = \k \ast \S + \n, \label{eq:model}   	
\end{equation}
where $\n$ is an additive noise, usually i.i.d.~Gaussian with
zero mean.  This model, while often satisfactory, does not take
into account the fact that blur due to defocus or rotational camera motion is not uniform~\cite{levin}. But, at least
{\em locally}, it is sufficient to describe many types of blurs.

In the noiseless case when the filter is a known {\em imperfect} low-pass
filter---that is, there is no zero in its Fourier transform, the blurring operator is
invertible and deblurring amounts to inverting the Fourier transform. 
However, noise is always present in natural images, and even a small amount 
dominates the signal in high frequencies, leading to numerous artefacts. 
Regularization methods have been extensively studied to tackle this problem~\cite{Hansen}. 
They usually impose smoothness constraints on the reconstructed images.
The most recent and effective algorithms  in this
line of work usually adopt a two-step approach~\cite{BM3D,foi2006,SVGSM}: first,
a simple regularized inversion of the blur is performed, 
then the resulting image is processed with classical
denoising algorithms to remove artefacts.
Various denoising methods have been used for this task: for instance, a Gaussian scale mixture model (GSM)~\cite{SVGSM},
the shape-adaptive discrete cosine
transform~\cite{foi2006}, or block matching with 3D-filtering kernel regression~\cite{BM3D}.

The digital zooming literature has seen in recent years the development of another line of research, following the 
 exemplar-based method introduced by Freeman et al.~\cite{freeman}.
Correspondences between high-resolu\-tion patches and low-resolution ones are
learned by building a large database of such pairs. 
This idea has been successfully exploited by Glasner et al.~\cite{glasdner}, leading to
state-of-the-art results. Along the same line, but using sparse image representations 
instead, pairs of corresponding patches are used by Yang et al.~\cite{YiMa} to jointly
learn high and low-resolution dictionaries. As shown in
Section~\ref{sec:formulation}, the method we propose exploits these
exemplar-based ideas as well, but in a significantly different way.

\subsection{Learned Sparse Representations}
Like several recent approaches to image restoration \cite{elad,YiMa}, our
method is based on the sparse decomposition of image patches.  Using a
dictionary matrix $\D =[\d_1,\ldots,\d_k]$ in $\Real^{m \times k}$, a signal $\x$ in $\Real^m$ is reconstructed as a linear combination of a few columns of 
$\D$, called atoms or dictionary elements.
In typical image processing applications, $m$ is relatively small, for
instance $m=64$ for image patches of size  $8 \times 8$ pixels, and $k$ can be
larger than $m$, e.g., $k=256$.
We say that the dictionary $\D$ is well adapted to a vector $\x$ when
there exists a sparse vector $\alphab$ in $\Real^k$ such that $\x$ can be approximated by the product $\D\alphab$.

Exploiting these types of models usually requires a ``good'' dictionary.
It can either be prespecified or designed by adapting its content
to fit a given set of signal examples.  Choosing prespecified atoms is
appealing: The theoretical properties of the corresponding
dictionaries can often be analysed, and, in many cases, it leads to fast
algorithms for computing sparse representations.  This is indeed the
case for wavelets~\cite{mallat}, curvelets, steerable wavelet filters, short-time Fourier
transforms, etc.  The success of the corresponding dictionaries in
applications depends on how suitable they are to sparsely describe the
relevant signals. 

Another approach consists of learning the dictionary on a set of signal
examples. The sparse decomposition of a patch~$\x$ on a fixed dictionary $\D$
can be achieved by solving an optimization problem called Lasso in statistics~\cite{tibshirani}
or basis pursuit in signal processing~\cite{chen}:
\begin{equation}
     \min_{\alphab \in \Real^k} \normd{\x-\D\alphab}+\lambda \normu{\alphab},
\end{equation}
where the code $\alphab$ in $\Real^k$ is the representation of $\x$ over the dictionary
$\D$, and $\lambda$ is a parameter for controlling the sparsity of the
solution.\footnote{It is well known that $\ell_1$ regularization yields a
sparse solution for $\alphab$, but there is no direct analytic link between
the value of $\lambda$ and the corresponding effective sparsity that it
yields.} Following an idea originally introduced in the neuroscience community by Olshausen and Field~\cite{olshausen},
 Aharon et al.~\cite{elad} have empirically shown that learning a dictionary $\D$ 
adapted to natural images could lead to better performance for image denoising than using off-the-shelf ones.
  For a database of $n$ patches of size $m$, a
dictionary is learned by solving the following optimization problem
\begin{equation}
   \min_{\D \in \Drond, \alphab_i \in \Real^k} \frac{1}{n} \sum_{i=1}^n \|\x_i-\D\alphab_i\|_2^2 + \lambda \|\alphab_i\|_1 , \label{eq:dict}
\end{equation}
 where $\x_i$ is the $i$-th patch of the training set, and $\alphab_i$ is its
associated sparse code. To prevent the columns of $\D$ from being arbitrarily
large (which would lead to arbitrarily small values of $\alphab$), the dictionary~$\D$ is
constrained to belong to the set $\Drond$ of matrices in $\Real^{m \times k}$
whose columns have an $\ell_2$ norm less than or equal to one.

 Several algorithms have been designed to address this problem.  They either
update~$\D$ and the vectors~$\alphab_i$ in a sequential way~\cite{elad}, or are
based on stochastic approximations~\cite{mairal2,olshausen}.

\subsection{Deblurring with Dictionaries}
Several methods using dictionaries for deblurring have been presented in recent years \cite{YiMa,yu2010}. Yu et al.~\cite{yu2010}, while not learning 
a dictionary as presented in the previous section, uses orthogonal basis obtained with principal component analysis (PCA). By ``learning'' several such dictionaries (one for each edge direction), and by choosing the best dictionary
for each patch, the sharp patch can be reconstructed. 

In the pioneering work by Yang et al.~\cite{YiMa}, a pair of dictionaries ($\D_b$,$\D_s$) is used, one dictionary for preprocessed
 blurred patches and the other for sharp patches. The preprocessing consists in the concatenation of oriented high-pass filters (gradients and Laplacian filters). 
During training, $\D_b$ and $\D_s$ are learned for
representing simultaneously (with the same sparse code) the sharp patches with
$\D_s$ and the preprocessed blurred patches with $\D_b$.
At test time, given a new preprocessed blurry patch~$\x$, a sparse code $\alphab$ is obtained by decomposing $\x$ using~$\D_b$, and ones hopes $\D_s\alphab$ to be a good estimate of the unknown sharp patch.

This method, while appealing by its simplicity, suffers from an
asymmetry between training and testing: Whereas in the learning phase, both
blurred and sharp patches are used to obtain the sparse codes, at test time the code 
is only computed using the blurry patches.  
Our method addresses this problem by a different training formulation. Moreover preprocessing the data has empirically not shown to be necessary.
\section{Proposed Approach} \label{sec:formulation}
We show in this section how to learn dictionaries ada\-pted to the deblurring and digital zoom tasks.
As in exemplar-based methods \cite{freeman,glasdner,YiMa}, we are given a training set of $n$ pairs of patches (obtained from pairs of blurry/sharp
images), that are used to estimate model parameters. 
Unlike the classical dictionary learning problem of Eq.~(\ref{eq:dict}) which is
unsupervised, our deblurring and digital zoom formulation is therefore supervised, trying to predict the sharp patches from the
blurry ones.

To predict a sharp pixel value, it is necessary to observe neighbouring blurry pixels.
Sharp patches and blurry patches may therefore have different sizes, which we denote
respectively by $m_b$ and $m_s$, with $m_b$ larger than
$m_s$. 
During the test phase, we observe a test image $\B$ and try to estimate the
underlying sharp image $\S$ according to Eq.~(\ref{eq:model}),
assuming of course that its blur is of the same nature as the one used during the
training phase.
The following sections present different formulations to recover an
estimate of~$\S$.

\subsection{Linear Model}
Blurring is, at least locally, a linear operation resulting from the
convolution of a sharp image with a filter.
When the support of the blur kernel is small compared to the patch sizes $m_s$ and $m_b$,
one can assume a linear relation between the blurry and sharp patches.
Thus, a simple approach to the deblurring problem consists of
learning how to invert this linear transform with a simple ridge regression
model.

{\bf Training Step: }
A training set $(\b_i,\s_i)$, $i=1,\ldots,n$ of pairs of blurry/sharp patches is given.
The training step amounts to finding the matrix $\W$ in $\Real^{m_s \times m_b}$ that
solves the following optimization problem:
\begin{equation}
   \min_{\W \in \Real^{m_s \times m_b}} \frac{1}{n} \sum_{i=1}^{n} \|\s_{i}-\W\b_{i}\|_2^2 
 + \mu\normF{\W},
\end{equation}
where $\|\W\|_F$ denotes the Frobenius norm of the matrix~$\W$, $n$ is the
number of training pairs of patches, and~$\mu$ is a regularization parameter,
which prevents overfitting on the training set and ensures that the learning
problem is well posed.  When $n$ is very large (several millions), overfitting
is unlikely to occur and setting $\mu$ to a small value (e.g., $\mu=10^{-8}$ in
our experiments) leads to acceptable results in practice. For this reason,
and for simplifying the notation, we drop the term $\mu\normF{\W}$ in the rest of the paper.

{\bf Testing Step:} The parameters $\W$ are now fixed, and we are given a noisy test image $\B$, the goal being to recover a sharp
estimate $\S$.  However, as mentioned in Section~\ref{sec:related}, the noise dominates
the signal in high frequencies, and in practice the linear model, which
basically tries to invert the blur operator, leads to poor results despite the
large amount of training data. Improvements can be achieved using recent
denoising algorithms, either by pre-processing $\B$ to remove some of its
noise, and/or by post-processing the sharp estimate to remove artefacts.

We now pre-process $\B$ and call $\tilde{\B}$ its denoised version, which is
obtained with a denoising algorithm~\cite{mairalNonLocal},
and respectively denote by $\tilde{\b}_i$ and $\s_i$ the patches of $\tilde{\B}$
and $\S$ centered at the pixel $i$, using any indexing of the image pixels. Note that 
the patches $\s_i$ are here different from the ones in the training set, even though we
use for simplicity the same notation.  We assume with our learned linear model that the relation 
$\s_i \approx \W \tilde{\b}_i$ holds for the patch indexed by $i$.
According to this model, the problem of reconstructing the sharp image $\S$ can be written as:
\begin{equation}
   \min_{\S}  \frac{1}{n_s} \sum_{i=1}^{n_s} \|\s_i-\W \tilde{\b}_i\|_2^2,\label{eq:average}
\end{equation}
where $n_s$ is the number of patches in the image $\S$. By using such a local
linear model, and since the patches overlap, each pixel of the image $\S$
admits as many predictions as patches it belongs to. The solution of Eq.~(\ref{eq:average}) is the average of the different predictions at each pixel, which is a classical way of aggregating estimates in patch-based methods~\cite{elad}.

This model is easy to optimize and to understand but has several limitations.
First, small mistakes made during the denoising process can be amplified
by the deblurring step.

Second, when the blur kernel totally suppresses some of the high frequencies of
the image, putting them to zero, one cannot recover them with a local linear model: in the Fourier
domain it correspond to a multiplication of the nullified coefficient by a finite number. This is one of 
the motivations for introducing a nonlinear model based on
sparse representations to overcome these limitations.

\subsection{Dictionary Learning Formulation}
In a recent paper, Yang et al.~\cite{YiMa} have shown that learning multiple
dictionaries to establish correspondences between low- and high-resolution image
patches is an effective approach to digital zoom. Following this idea, we
propose to learn a pair of dictionaries~$\D_s$ in $\Real^{m_s \times k}$ and
$\D_b$ in $\Real^{m_b \times k}$ to reconstruct patterns that the
linear model presented in the previous section cannot recover.

{\bf Training step:}
Given again a training set $(\b_i,\s_i)$, $i=1,\ldots,n$ of pairs of blurry-noisy/sharp patches,
we address
\begin{equation}
   \min_{\D_b\in \Drond,\D_s,\W} \frac{1}{n}\sum_{i=1}^{n} \|\s_{i}-\W{\tilde{\b}}_{i}-\D_s\alphab^\star(\b_i,\D_b)\|_2^2,\label{eq:dict2}
\end{equation}
where $\alphab^\star(\b_i,\D_b)$ is the solution of the following sparse coding problem
\begin{equation}
   \alphab^\star(\b_i,\D_b) \defin \argmin_{\alphab \in \Real^k} \normd{\b_i -\D_b\alphab}+\lambda \normu{\alphab},\label{eq:sparsecoding}
\end{equation}
which is unique and well defined under a few reasonable assumptions on the
dictionary $\D_b$~(see \cite{mairal2} and references therein for more details).\footnote{
We have empirically found for our deblurring and super-resolution tasks on natural image 
patches and our dictionaries that the solution of Eq.(7) was always unique. For different
 tasks or data, the possible non-uniqueness of the Lasso  solution could be an issue (see \cite{mairalPAMI}).}
The patch $\tilde{\b}_i$ is a denoisied version of $\b_i$. The matrices $\D_b$ and~$\D_s$ are two dictionaries jointly learned such that
for all $i$, $\W\tilde{\b_i}+\D_s\alphab^\star(\b_i,\D_b)$ is a good
estimator of the sharp patch $\s_i$. Summing two different predictors is a classical way
of combining two models. In this case, we are hoping that the addition of the dictionary term to 
the linear term will permit a better recovery of the high frequencies. The two models 
are optimized jointly and are not just an averaging of two independent predictors. 

Note that $\D_s$ does not need to be regularized in our formulation. We indeed
assume that a large amount of training data is available and as a consequence
our model does not suffer from overfitting. 

{\bf Testing step:}
According to our model, and using the same notations as in
Eq.~(\ref{eq:average}), our estimate $\hat{\S}$ at test time is achieved by
solving the following optimization problem
\begin{equation}
   \min_{\S}  \frac{1}{n_s} \sum_{i=1}^{n_s} \|\s_i-\W \tilde{\b}_i-\D_s\alphab^\star(\b_i,\D_b)\|_2^2,\label{eq:average2}
\end{equation}
where $\s_i, \b_i, \tilde{\b_i}$ are respectively here the patches centered at
pixel $i$ of the sharp image $\S$, the blurry, noisy image $\B$ and the blurry, denoisied image $\tilde{\B}$.

The optimization problem defined in Eq.~(\ref{eq:dict2}) is harder than the
classical dictionary learning of Eq.~(\ref{eq:dict}) or the one formulated by
Yang et al.~\cite{YiMa}, but this formulation presents advantages.  

In the work of Yang et al.~\cite{YiMa}, the sparse coefficients~$\alphab$ are
obtained during the training phase by jointly decomposing blurry patches $\b_i$
and sharp patches~$\s_i$ onto two learned dictionaries $\D_b$ and $\D_s$.
Such a model aims to ensure that there always exists a sparse code
$\alphab$ that both fits the patches~$\b_i$ and~$\s_i$. However, at test time, since
the sharp patches are not available, the vectors $\alphab$ can only be computed
from blurry patches $\b_i$, and the fact that the resulting~$\alphab$ should be good for the 
corresponding sharp patch $\s_i$ is not guaranteed.

Our approach does not suffer from this issue since the sparse coefficients
$\alphab$ are always obtained on blurry patches only, both during the training and testing 
phase. We learn the dictionaries $\D_b$ and $\D_s$ and the linear predictor
$\W$ such that~$\s_i$ is well predicted \emph{given} a patch $\b_i$.
Whereas this solves the issue mentioned above, it leads to more 
challenging optimization problems than \cite{YiMa}.
The optimization method we propose builds upon \cite{mairalPAMI}, which provides
a general framework for solving such dictionary learning problems.
The method is presented briefly in Section~\ref{sec:optim}.

We have presented so far a framework adapted to the deblurring
task, where we wanted to obtain a sharp image from a blurry one. The problem of
digital zoom consists of increasing the resolution of an image, but  can be
formulated as a deblurring problem in a simple way:  A low-resolution image can
indeed be turned into a blurry high-resolution image with any interpolation
technique, the task of digital zoom being then to \emph{deblur} this new
image. The training pairs of images can be
generated by downsampling high-resolution images. Note that the antialiasing
filter applied during downsampling and the choice of the interpolation
method are important. We worked with the antialiasing from
the Matlab function \emph{imresize}.

\section{Optimization} \label{sec:optim}
The formulation of Eq~(\ref{eq:dict2}) for learning a pair of dictionaries
$\D_b$ and $\D_s$ and a linear predictor $\W$ for the deblurring task is a large-scale learning problem,
where many training samples $(\b_i,\s_i)$ can easily be available. The main
difficulty in the optimization comes from the terms $\alphab^\star(\b_i,\D_b)$,
which are defined as solutions to the sparse coding problem of
Eq.~(\ref{eq:sparsecoding}). The vectors $\alphab^\star(\b_i,\D_b)$ therefore depend
 on the dictionary $\D_b$ and are not differentiable with respect to
it, preventing us from using a direct gradient descent method.

However, despite these two drawbacks, it has been shown in \cite{mairalPAMI}
that such problems enjoy a few asymptotic properties that make it possible to
use stochastic gradient descent when the number of training samples is large.
Assuming an infinite
training set $(\b_i,\s_i)$ that are i.i.d.~samples drawn from some probability distribution,
and under mild assumptions, we define the asymptotic cost function 
\begin{equation}
\begin{split}
f(\D_b,\D_s,\!\W) \!&\defin\!\! \lim_{n \to +\infty} \! \frac{1}{n} \! \sum_{i=1}^{n} \! \|\s_{i}\!-\!\W{\b}_{i}\!-\!\D_s\alphab^\star(\b_i,\D_b)\|_2^2, \\
                   & = \E_{(\b,\s)}\big[\|\s-\W\b-\D_s\alphab^\star(\b,\D_b)\|_2^2\big],
\end{split}
\end{equation}
where $(\b,\s)$ are random variables distributed according to the joint probability distribution of low/high-resolution patches.

The optimization of cost functions that have the form of an expectation over a
supposedly infinite training set is usually tackled with stochastic gradient
techniques (see \cite{mairalPAMI,mairal2} and references therein), that are iterative
procedures drawing randomly one element of the training set at a time.
Of course training sets are in practice finite, but we have empirically obtained
good results by optimizing on a large training set of $10$ millions of training patches.
This is indeed the approach proposed in \cite{mairalPAMI} for such problems, 
from which the following proposition can be derived.

\begin{proposition}{ { \bf [Differentiability of $f$]}}
   Assume that the training data $(\b,\s)$ admits a continuous probability density,
   and assume the same hypotheses on the dictionary $\D_b$ as in  \cite{mairalPAMI}.
   Then, $f$ is differentiable and
   \begin{equation}
       \begin{split}
          \nabla_\W f & = - \E_{(\b,\s)}[2(\s-\D_s\alphab^\star-\W\b)\b^T],\\
          \nabla_{\D_s} f & =  -\E_{(\b,\s)}[2(\s-\D_s\alphab^\star-\W\b)\alphab^{\star T}],\\
          \nabla_{\D_b} f & = - \E_{(\b,\s)}[2(\b\betab^{\star T}-\D_b\alphab^\star\betab^{\star T} - \D_b\betab^\star \alphab^{\star T})], \\
       \end{split} \label{eq:gradients}
   \end{equation}
   where $\alphab^\star$ denotes $\alphab^\star(\b,\D_b)$, and 
   \begin{equation}
      \betab^\star_{\Lambda^C}  = 0 ~\text{and}~
      \betab^\star_{\Lambda}  = -(\D_{b\Lambda}^T \D_{b\Lambda})^{-1}\D_{s\Lambda}^T(\s-\D_s\alphab^\star-\W\b), 
      \label{eq:beta}
   \end{equation}
   where $\Lambda$ denotes the indices of the nonzero coefficients of~$\alphab^\star$, for any vector $\u$, the vector $\u_\Lambda$
   contains the values of the vector $\u$ corresponding to the indices $\Lambda$, and for any matrix $\U$, the matrix $\U_\Lambda$ contains
   the columns of $\U$ corresponding to the indices $\Lambda$.
\end{proposition}

Algorithm 1 presents our method for learning $\D_s,\W$ and $\D_b$.
It is a stochastic gradient descent algorithm, which adapts~\cite{mairalPAMI} to our formulation. It draws randomly one element
of the training set at each iteration, computes the terms inside the
expectations of Eq.~(\ref{eq:gradients}), and moves the parameters
$\D_s,\W,\D_b$ one step in these directions.

Since $\D_b$ is constrained to be in the set $\DD$ defined in Eq.~(\ref{eq:dict}), 
an orthogonal projection on this set is required at each iteration of the algorithm. It
is denoted by $\Pi_\DD$.

\begin{algorithm}[hbtp]
   \caption{Dictionary Learning for Deblurring and Digital Zoom}
   \label{algo:sgd}
   \begin{algorithmic}
      \REQUIRE $(\b_i,\s_i)$, $i=1,\ldots,n$ (training set);
      $\lambda,\mu \in \Real$ (parameters);
      $\D_b \in \DD$ (initial ``blurry'' dictionary), $\D_s$ (initial ``sharp'' dictionary); $T$ (number of iterations); $t_0,\rho$ (learning rate parameters for the stochastic gradient descent).
      \FOR {$t= 1$ to $T$}
      \STATE Draw $(\b_t,\s_t)$ from the training set.
      \STATE Sparse coding: compute $\alphab^\star \defin \alphab^\star(\b_t,\D_b)$.
      \STATE Compute the active set: $\Lambda \leftarrow \{ j : \alphab^\star[j] \neq 0 \}$.
      \STATE Compute $\betab^\star$ according to Eq.~(\ref{eq:beta}).
      \STATE Choose the learning rate $\rho_t \leftarrow \frac{\rho}{t+t_0}$.
      \STATE Update parameters:
      \begin{displaymath}
         \begin{split}
         \W & \leftarrow  \W + \rho_t (\s_t-\D_s\alphab^\star-\W\b_t)\b_t^T,\\
         \D_s &\leftarrow \D_s + \rho_t (\s_t-\D_s\alphab^\star-\W\b_t)\alphab^{\star T},\\
          \D_b & \leftarrow \Pi_\DD \Big[ \D_b + \rho_t \big(\b\betab^{\star T}-\D_b\alphab^\star\betab^{\star T} - \D_b\betab^\star \alphab^{\star T}\big) \Big]. \\
       \end{split} \label{eq:update_param}
      \end{displaymath}
   \ENDFOR
   \RETURN $(\D_b,\D_s,\W)$ (learned model parameters).
\end{algorithmic}
\end{algorithm}

To improve the efficiency of the algorithm, we use a classical heuristic often referred to as : Instead of drawing a single pair of the training set at
the same time, we draw $\eta$ of them, e.g., $\eta=500$, compute $\eta$
directions given by Eq.~(\ref{eq:gradients}), and move the model parameters
$\D_b,\D_s,\W$ in the average direction. This improves the stability of the
stochastic gradient descent algorithm, and experimentally gives a faster
convergence.  Since our optimization problem is not convex, it requires a good
initialization.  We proceed as follows: (i) We learn a dictionary $\D_b$ 
using the unsupervised formulation of Eq.~(\ref{eq:dict}) with the software\footnote{The SPAMS toolbox 
is an open-source software available at: \textit{http://www.di.ens.fr/willow/SPAMS/} }
accompanying \cite{mairal2} on the set of patches $\b_i$.
(ii) We fix $\D_b$ and optimize Eq.~(\ref{eq:dict2}) with respect to $\W$ and $\D_s$, 
which is a convex optimization problem. In experiments, this procedure
provides us with a good initialization.

\section{Experiments} \label{sec:exp}
We present here experimental results obtained with our method and comparisons
with state-of-the-art methods. In all our experiments, after an initialization
step described in the previous section, we use the stochastic gradient descent
algorithm with one pass over a database of approximately $10$ millions of
training patches, which are extracted from a set of natural images.
All the images from this dataset are unrelated with the images used for testing
our method. Our implementation is coded in C++ and Matlab. 
Learning a dictionary takes usually a few hours on a recent computer, while
testing an image is faster (less than one minute for most of our test images).

\subsection{Non-Blind Deblurring with Isotropic Kernels}
To compare our method for the non-blind deblurring task, we have chosen a
classical set of images and types of blurs, which has been used in
several recent image processing papers~(see \cite{yu2010} and references therein).
Even though addressing such a synthetic non-blind deblurring task 
of course slightly deviates from real restoration problems with digital
cameras, it is still an active topic in the image processing community, and has
in fact proven useful in the past, leading to high-impact applications in
astronomic imaging~\cite{starck} for example (see Section~\ref{sec:astronomy}).

The different combinations of blurs and noises are detailed in Table~\ref{table:paramBlur}, with the shape of the 
blur kernel and the variance of the noise (which is Gaussian and white). They are used in other papers and 
go from strong-blur/weak-noise to weak-blur/strong-noise cases.

\begin{table}
\label{table:paramBlur} 
\caption{Experiments settings for the non-blind deblurring.}
\begin{tabular}{| c | l | c|}
\hline
\hline
Exp. & Blur kernel $\k$  & noise $\sigma^2$\\
\hline
\hline
1 & 9 x 9 \ uniform blur &  $0.308$ \\
\hline
2 &  \  $\k(x_1,x_2)=1/(1+x_1^2+x_2^2)$ &  $2$ \\
\hline
3 & \ $\k(x_1,x_2)=1/(1+x_1^2+x_2^2)$ &  $8$ \\
\hline
4 &  \ $\k=[1\ 4\ 6\ 4\ 1]^T[1\ 4\ 6\ 4\ 1] /256$ &  $49$ \\
\hline
5 & Gaussian blur of variance $\sigma_b=1$ & 25 \\
\hline
6 & Gaussian blur of variance $\sigma_b=2$ & 25 \\
\hline
\hline
\end{tabular}
\end{table}

\begin{figure*}
    \includegraphics[width=0.32\linewidth]{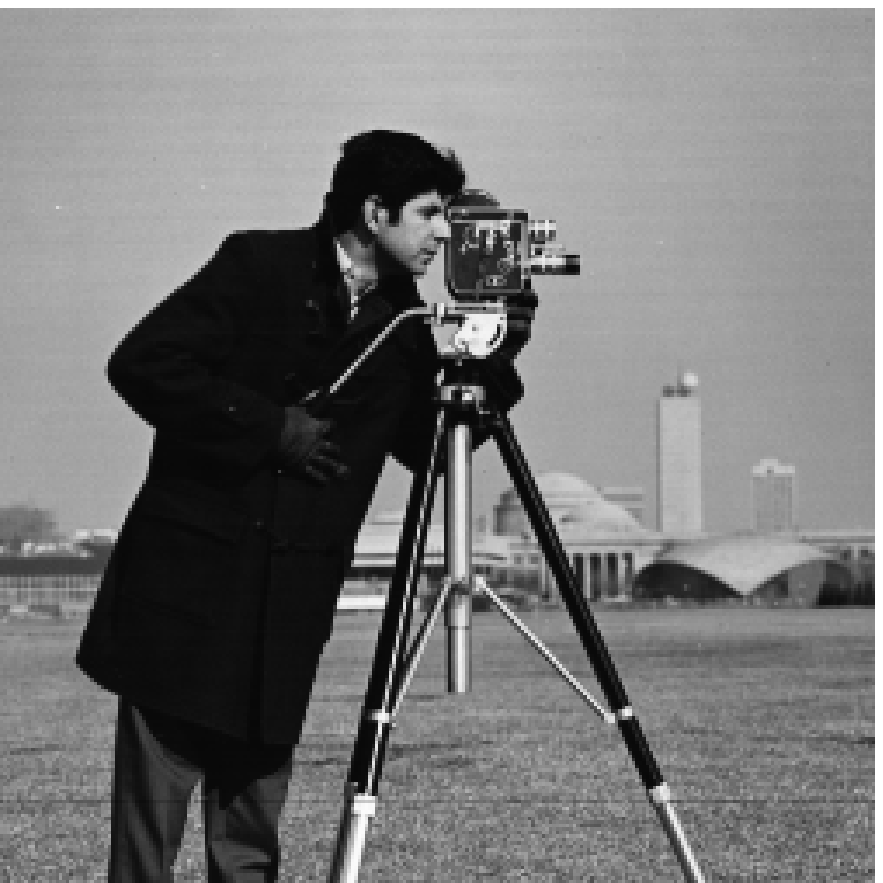} \hfill
    \includegraphics[width=0.32\linewidth]{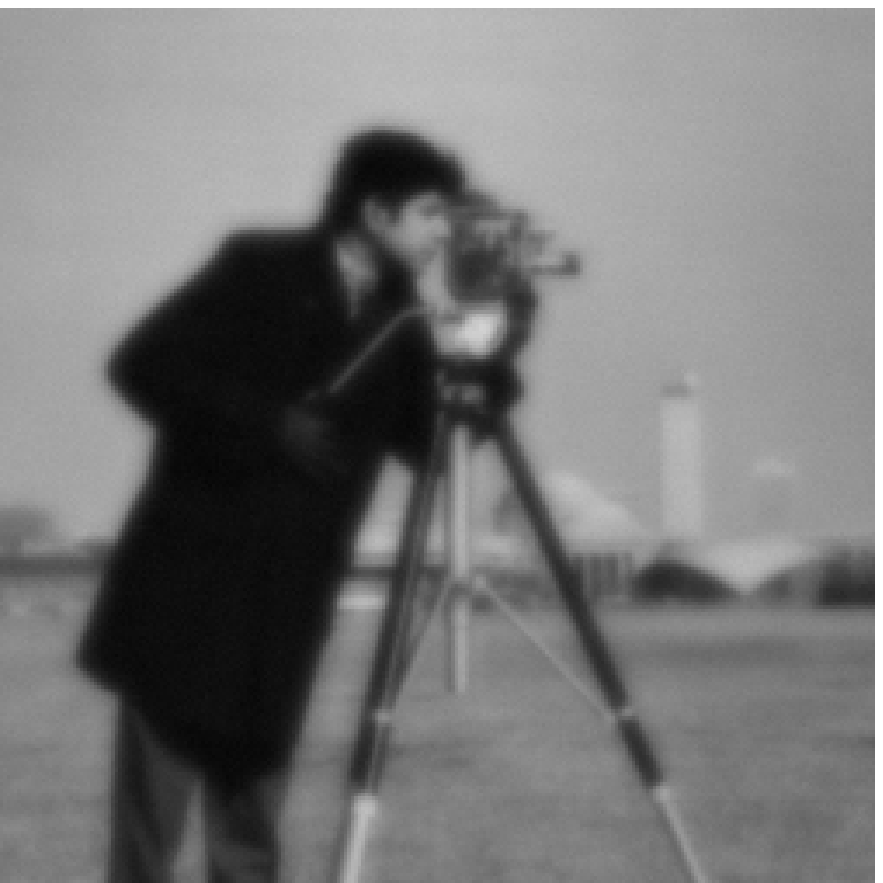} \hfill
    \includegraphics[width=0.32\linewidth]{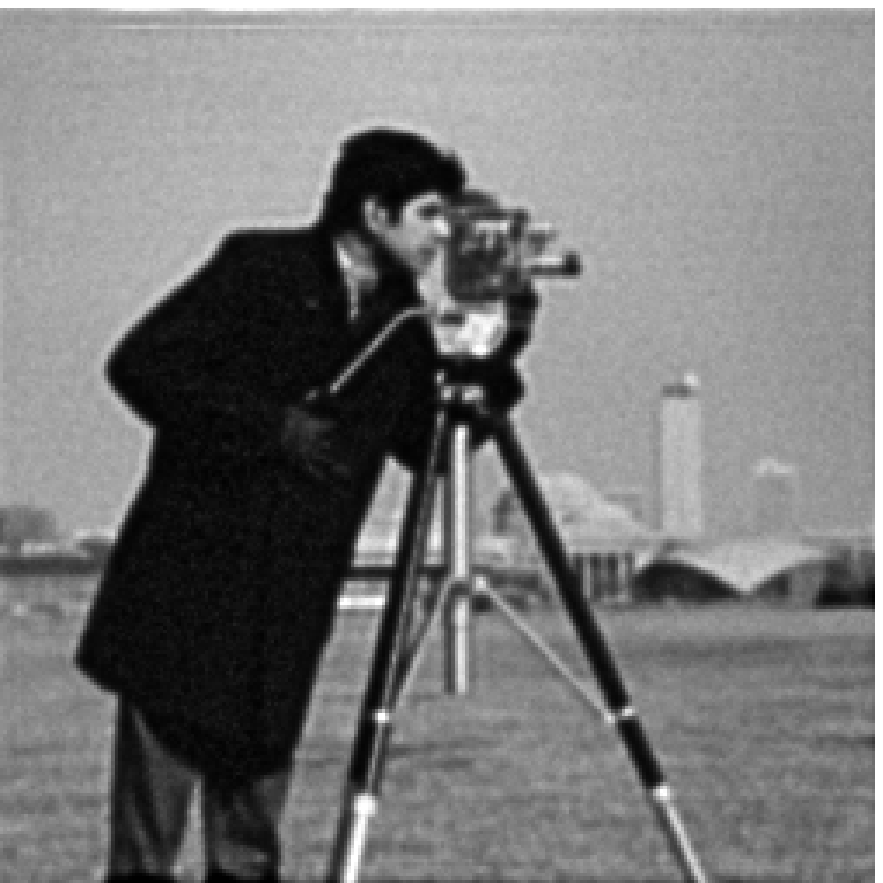} \\ \vfill
    \includegraphics[width=0.32\linewidth]{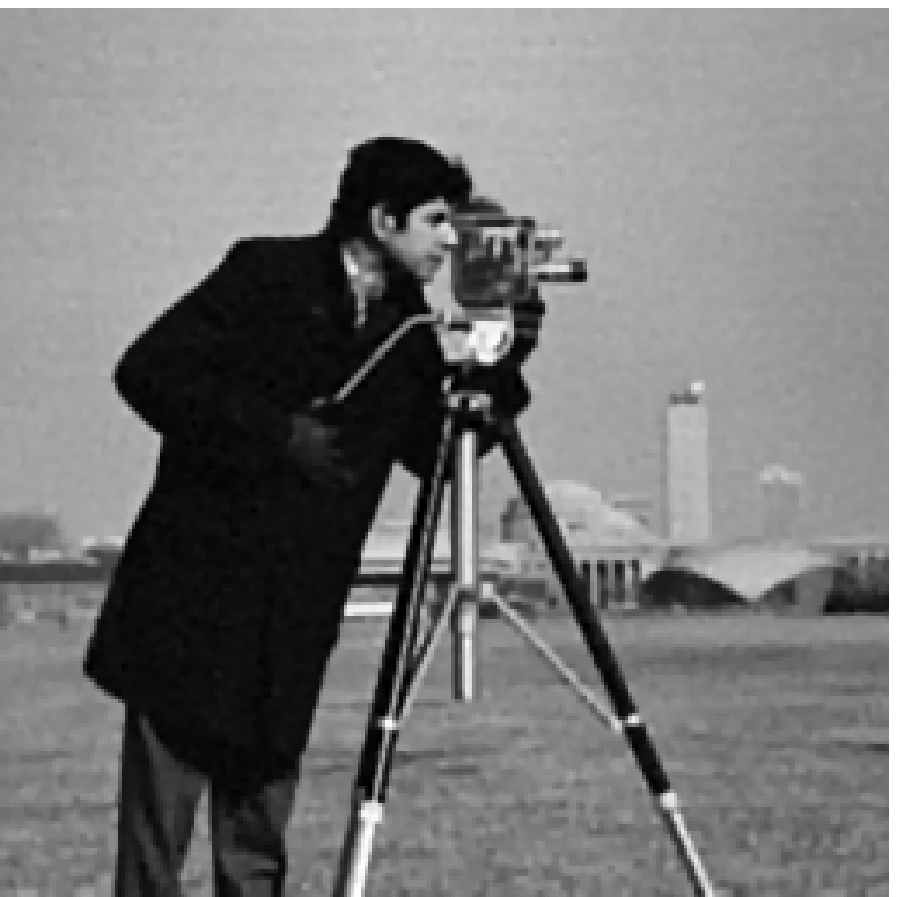} \hfill
    \includegraphics[width=0.32\linewidth]{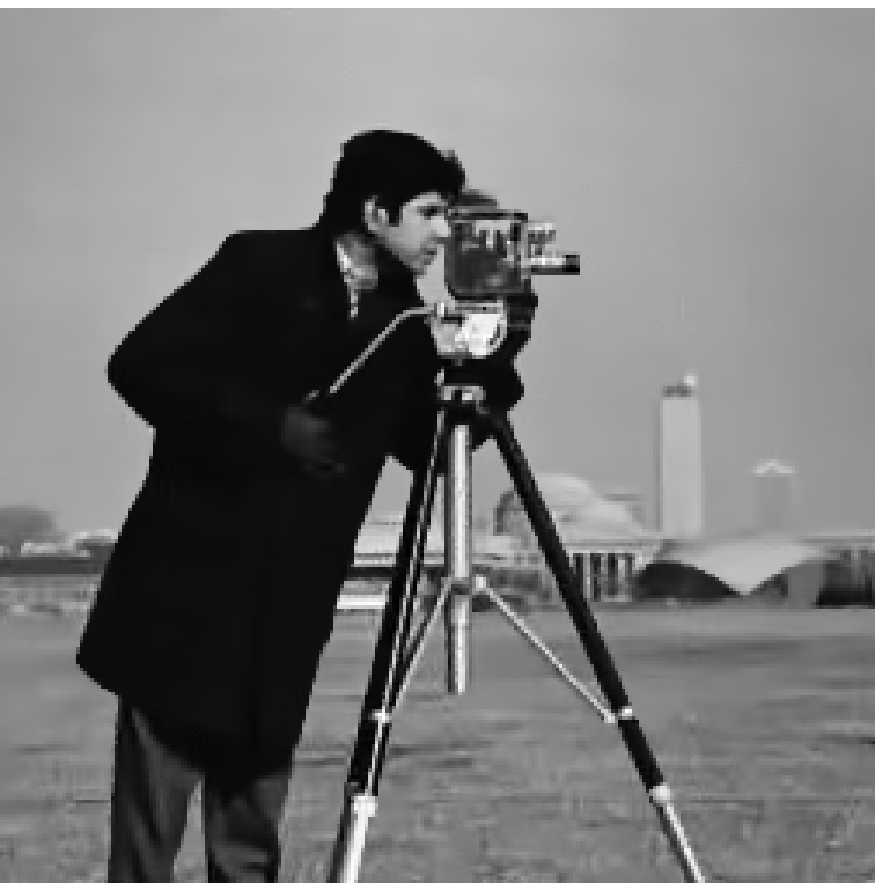} \hfill
    \includegraphics[width=0.32\linewidth]{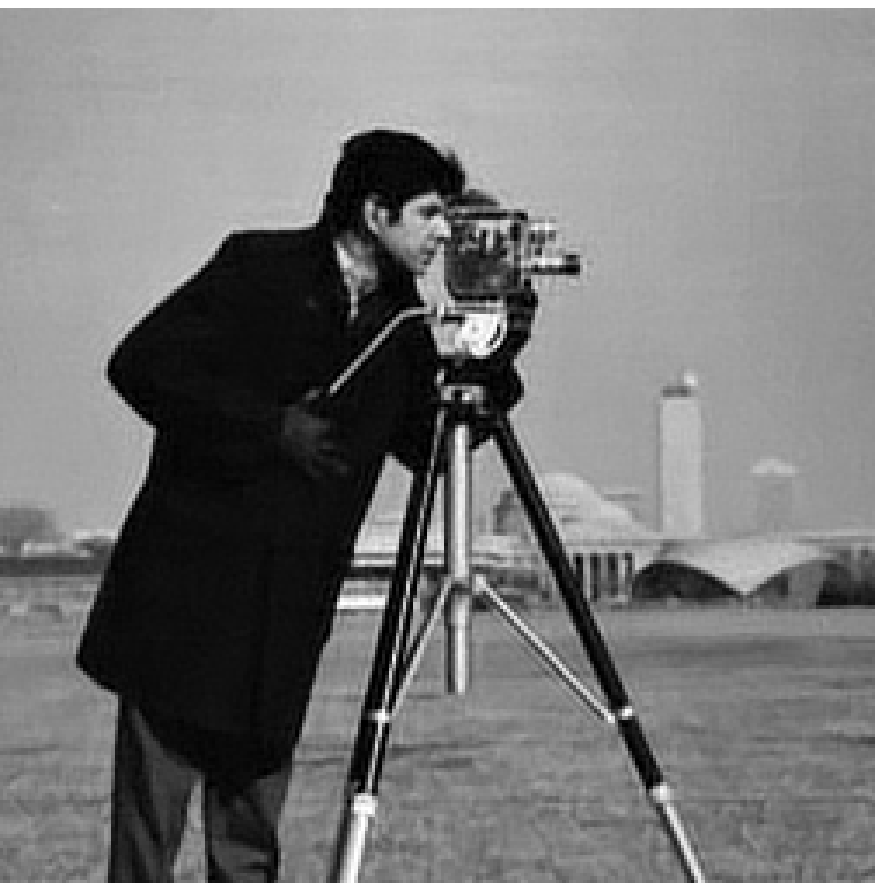}\\ \vfill
    \includegraphics[width=0.16\linewidth]{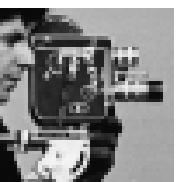} \hfill
    \includegraphics[width=0.16\linewidth]{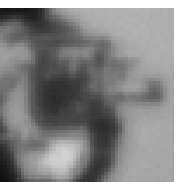} \hfill
    \includegraphics[width=0.16\linewidth]{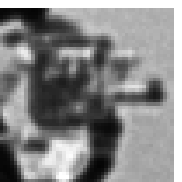} \hfill
    \includegraphics[width=0.16\linewidth]{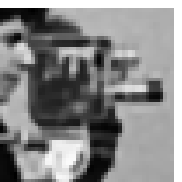} \hfill
    \includegraphics[width=0.16\linewidth]{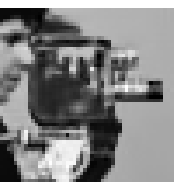} \hfill
    \includegraphics[width=0.16\linewidth]{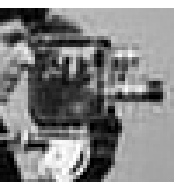} \\ \vfill
    \includegraphics[width=0.16\linewidth]{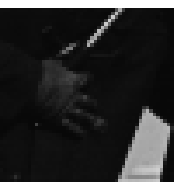} \hfill
    \includegraphics[width=0.16\linewidth]{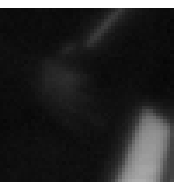} \hfill
    \includegraphics[width=0.16\linewidth]{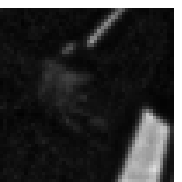} \hfill
    \includegraphics[width=0.16\linewidth]{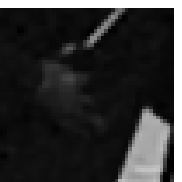} \hfill
    \includegraphics[width=0.16\linewidth]{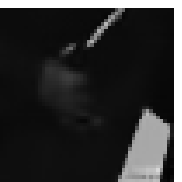} \hfill
    \includegraphics[width=0.16\linewidth]{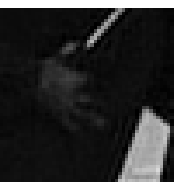}

  \caption{Examples of deblurring and close-up for the case 2. First two lines, from top to bottom, left to right: original image, blurry image,
 Richardson-Lucy, sparse gradient \cite{sparseGradient}, SA-DCT \cite{foi2006}, our method. 
Last two lines: close-ups in the same order. Best seen by zooming on a computer screen.}
  \label{fig:blur2}
\end{figure*}

\begin{figure*}
    \includegraphics[width=0.32\linewidth]{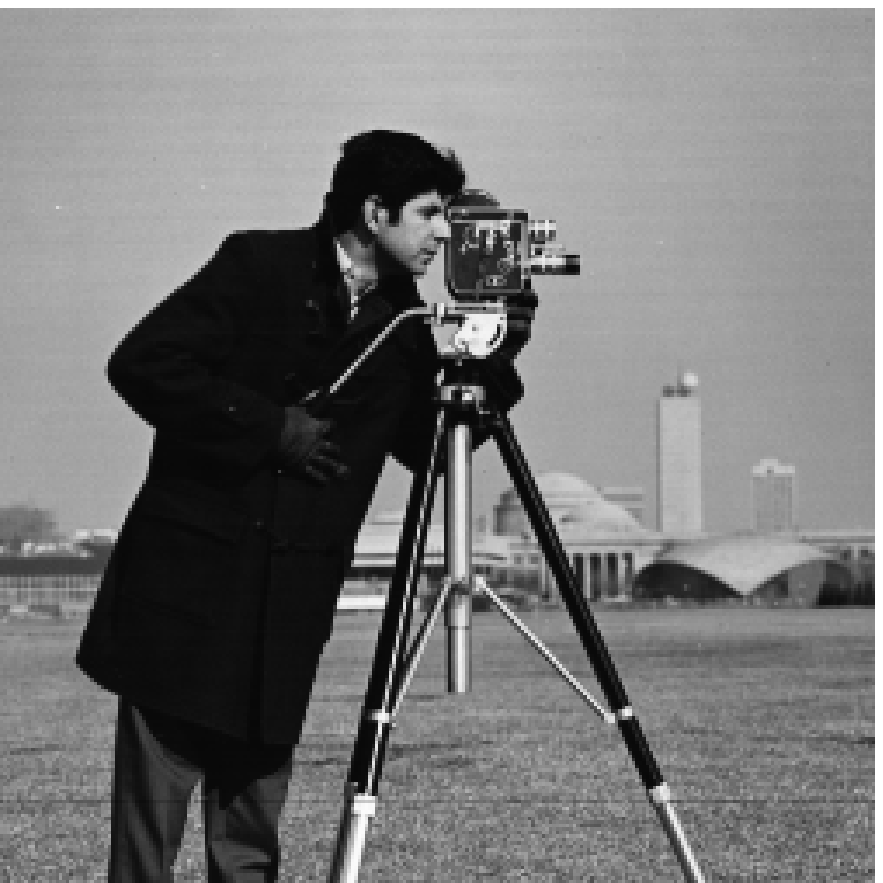} \hfill
    \includegraphics[width=0.32\linewidth]{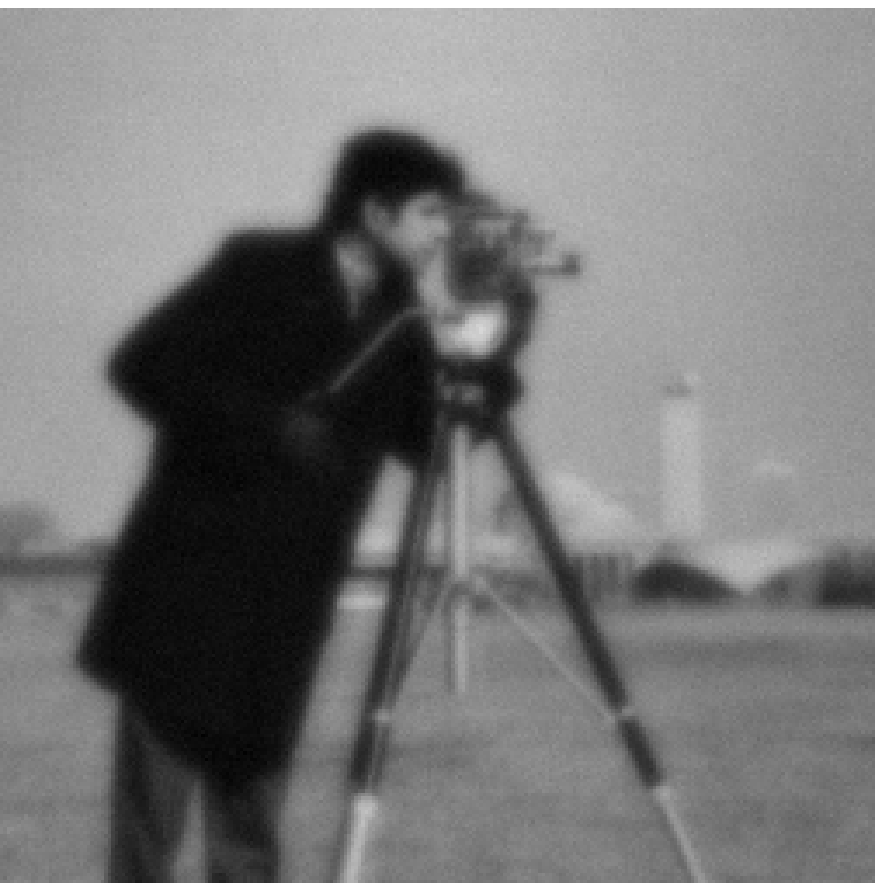} \hfill
    \includegraphics[width=0.32\linewidth]{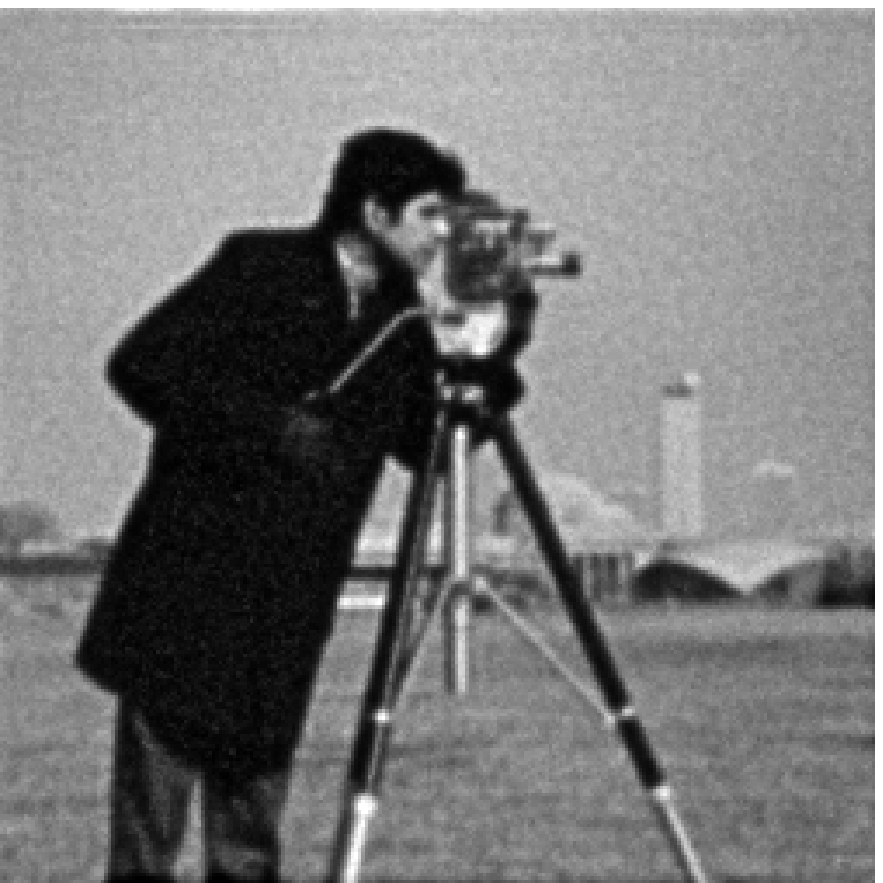} \\ \vfill
    \includegraphics[width=0.32\linewidth]{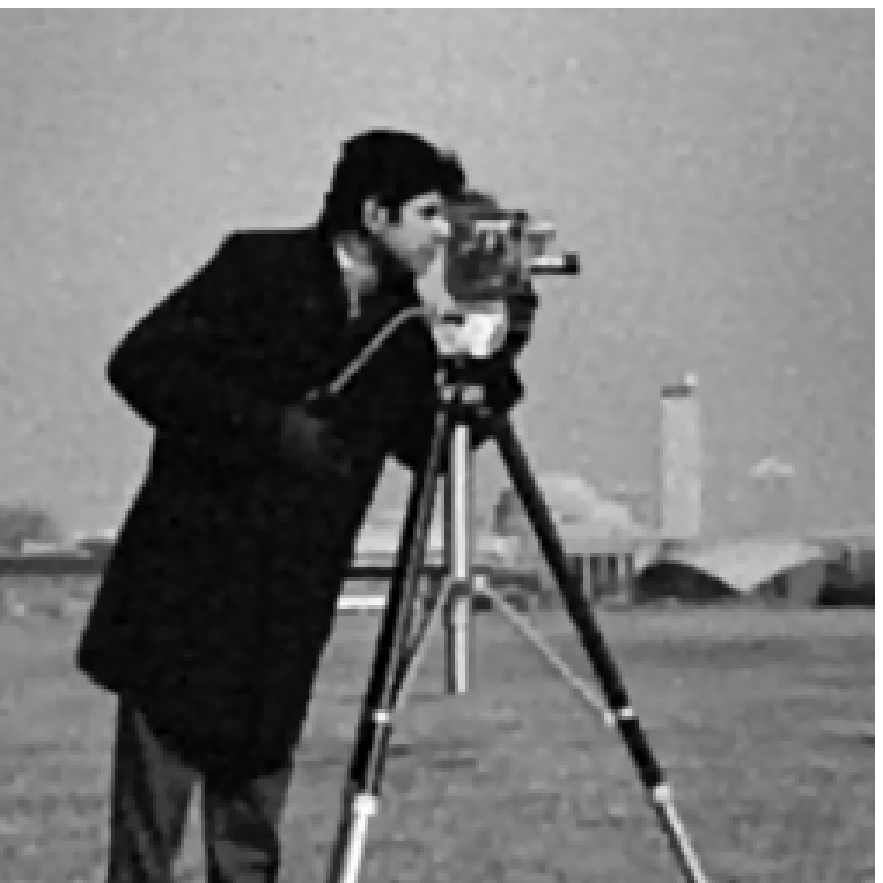} \hfill
    \includegraphics[width=0.32\linewidth]{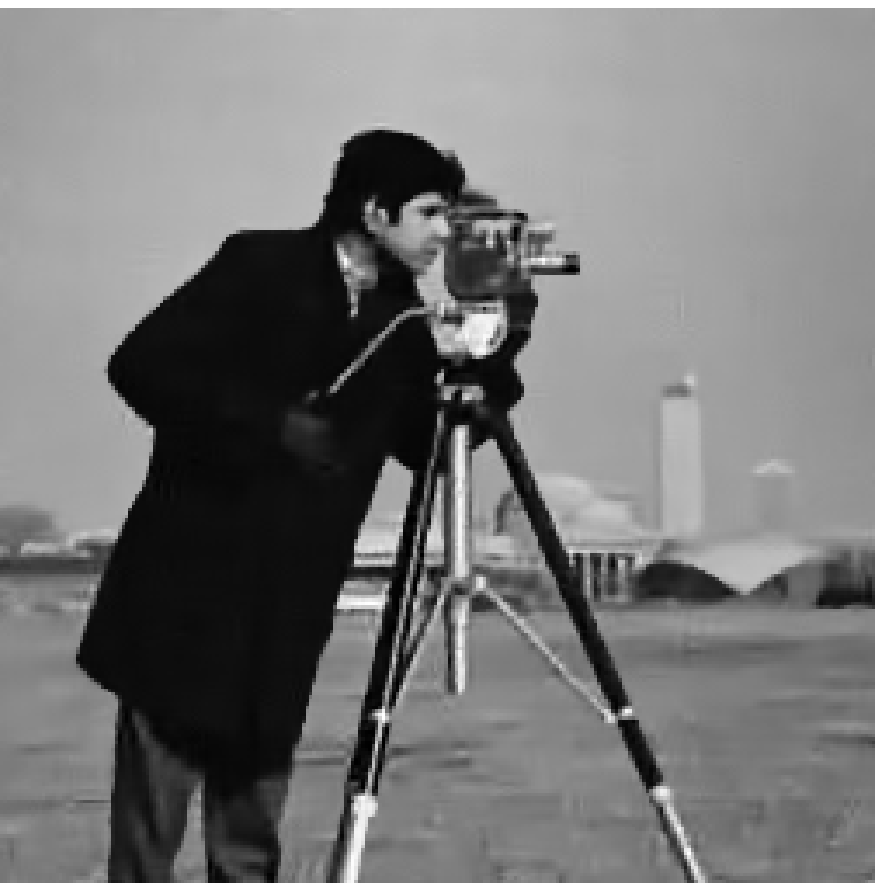} \hfill
    \includegraphics[width=0.32\linewidth]{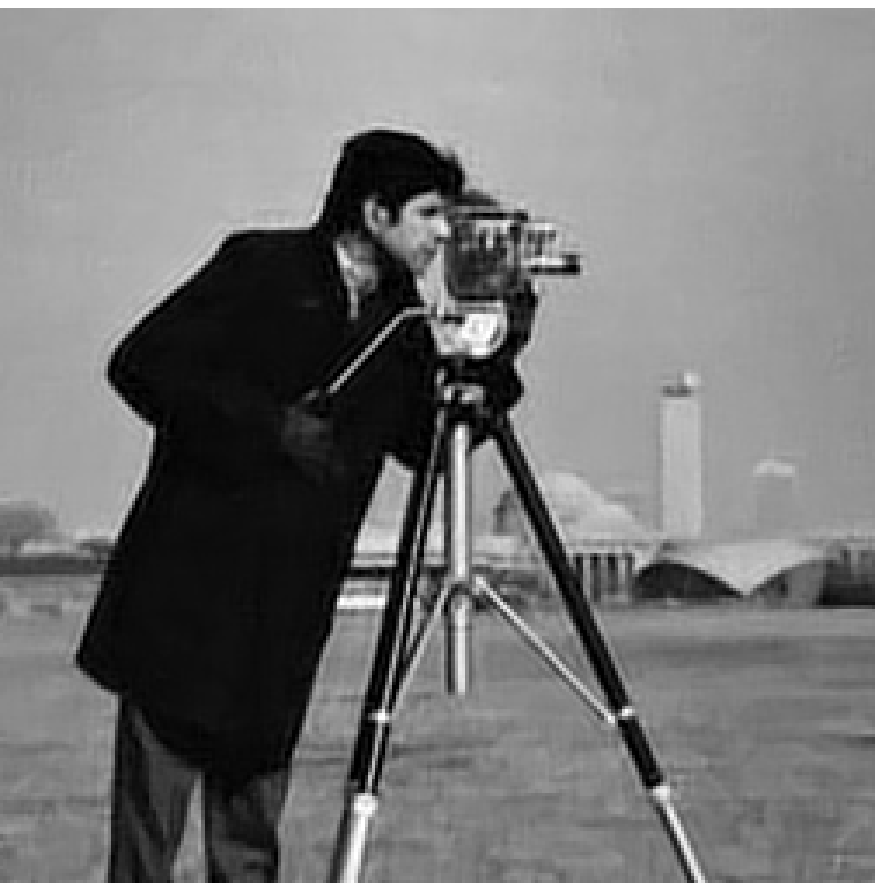} \\ \vfill
    \includegraphics[width=0.16\linewidth]{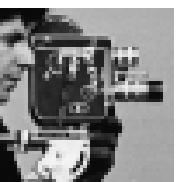} \hfill
    \includegraphics[width=0.16\linewidth]{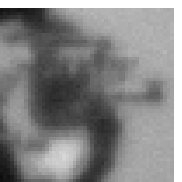} \hfill
    \includegraphics[width=0.16\linewidth]{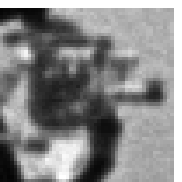} \hfill
    \includegraphics[width=0.16\linewidth]{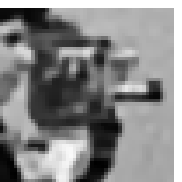} \hfill
    \includegraphics[width=0.16\linewidth]{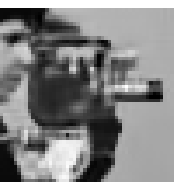} \hfill
    \includegraphics[width=0.16\linewidth]{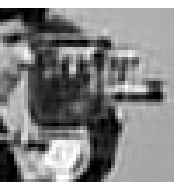} \\ \vfill
    \includegraphics[width=0.16\linewidth]{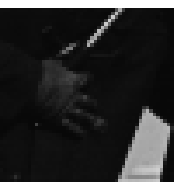} \hfill
    \includegraphics[width=0.16\linewidth]{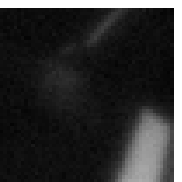} \hfill
    \includegraphics[width=0.16\linewidth]{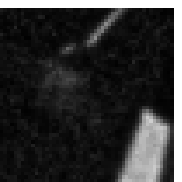} \hfill
    \includegraphics[width=0.16\linewidth]{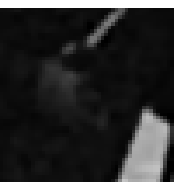} \hfill
    \includegraphics[width=0.16\linewidth]{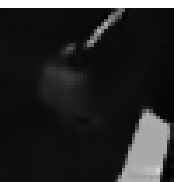} \hfill
    \includegraphics[width=0.16\linewidth]{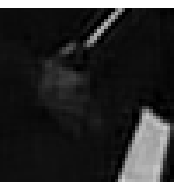}
    
  \caption{Examples of deblurring for the case 3. First two line, from top to bottom, left to right: original image, blurry image, Richardson-Lucy, sparse gradient \cite{sparseGradient}, SA-DCT \cite{foi2006}, our method. 
Last two lines: close-ups in the same order. Best seen by zooming on a computer screen.}
  \label{fig:blur3}
\end{figure*}

\begin{figure*}
    \hfill \includegraphics[width=0.49\linewidth]{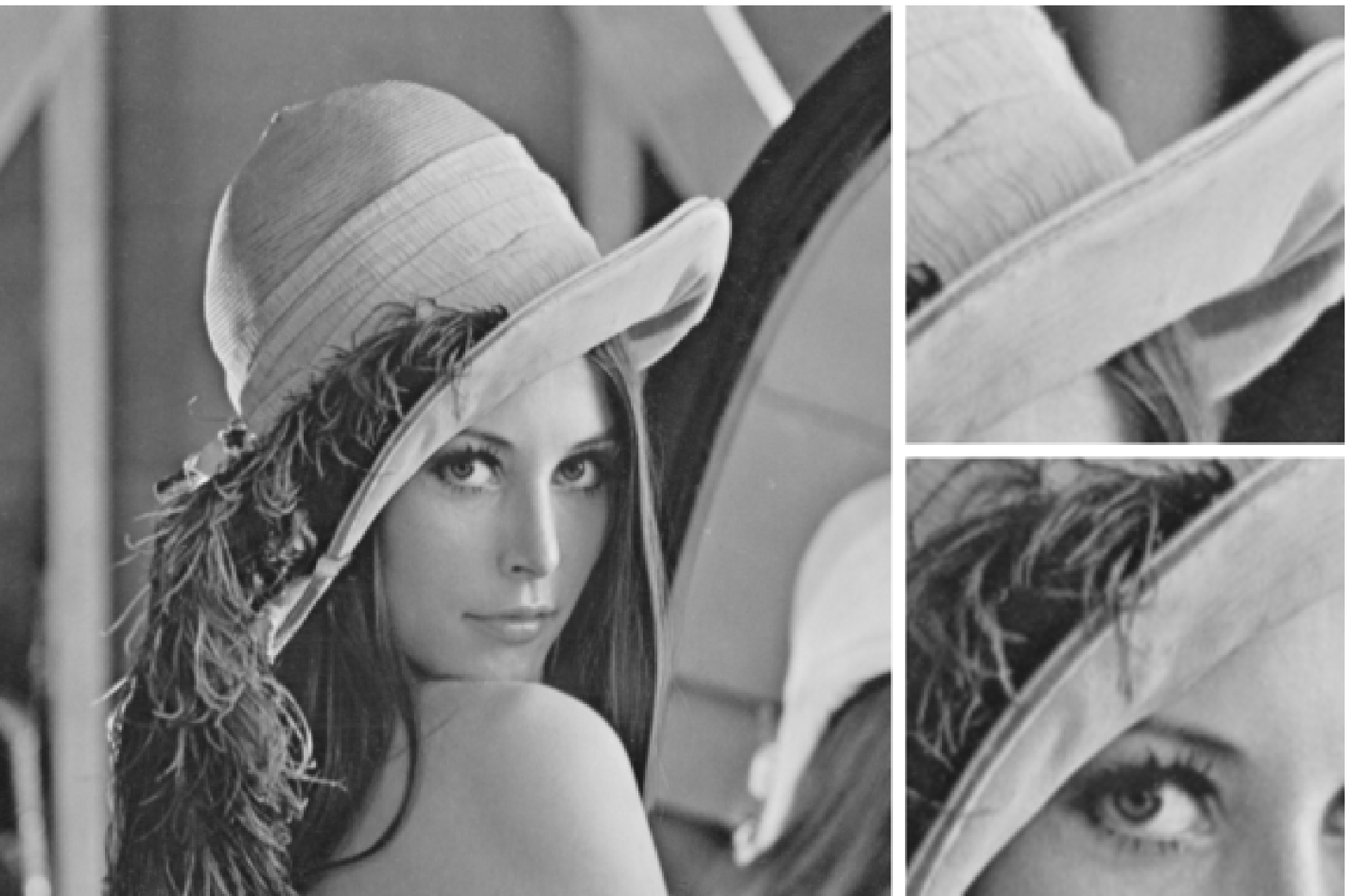} \hfill    
    \includegraphics[width=0.49\linewidth]{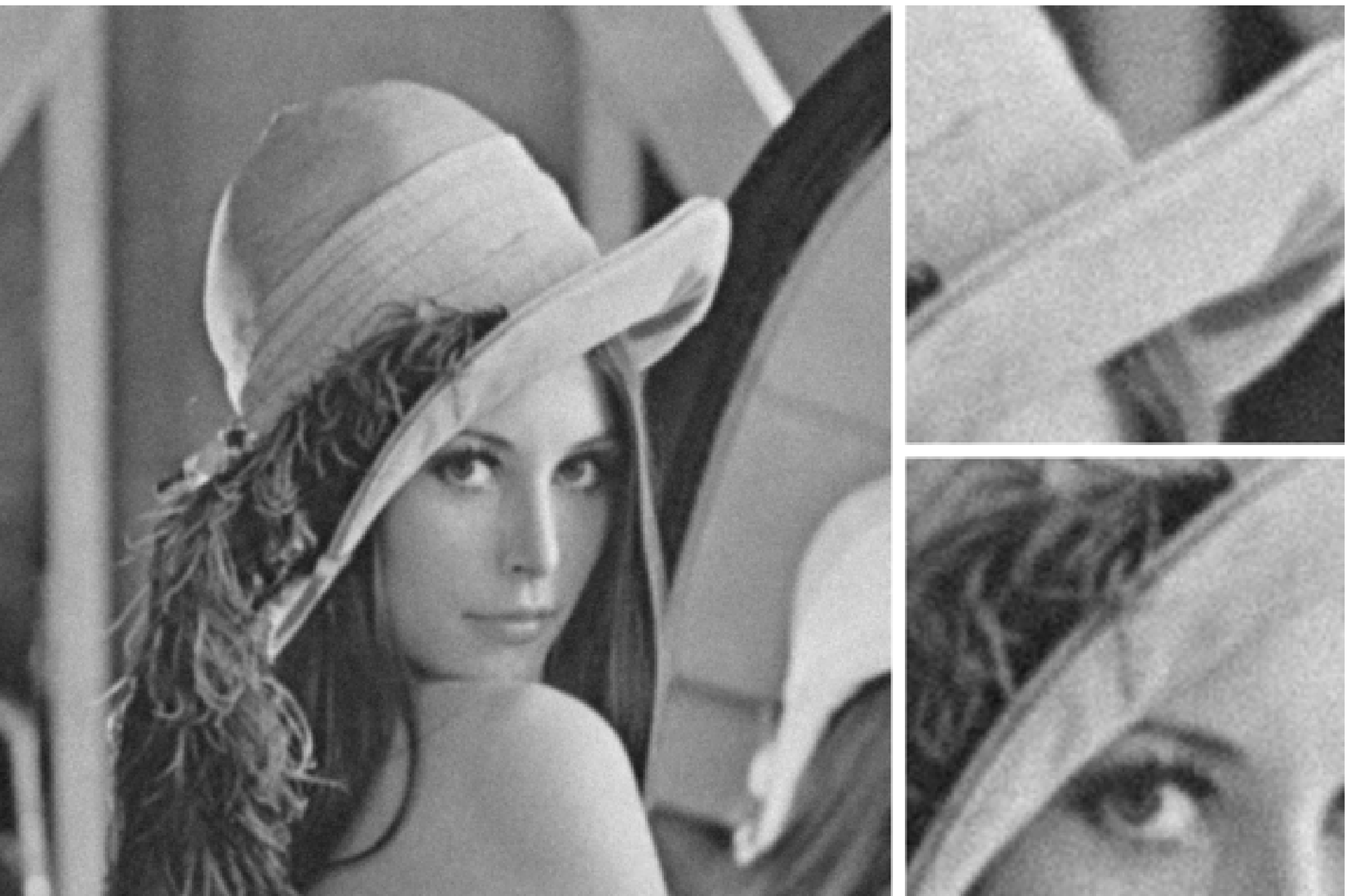} \hfill
\\ \vfill
    \hfill \includegraphics[width=0.49\linewidth]{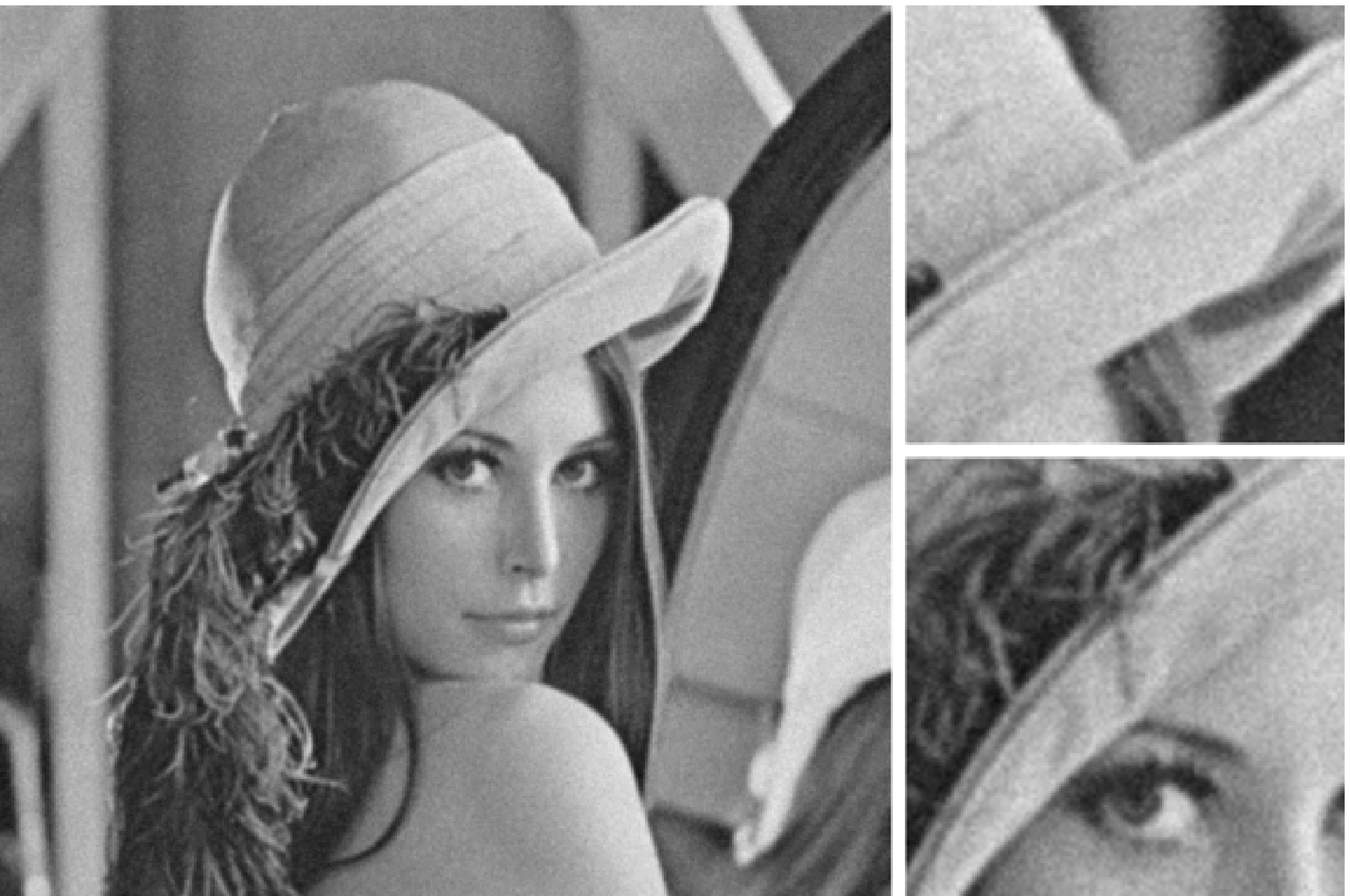} \hfill
    \includegraphics[width=0.49\linewidth]{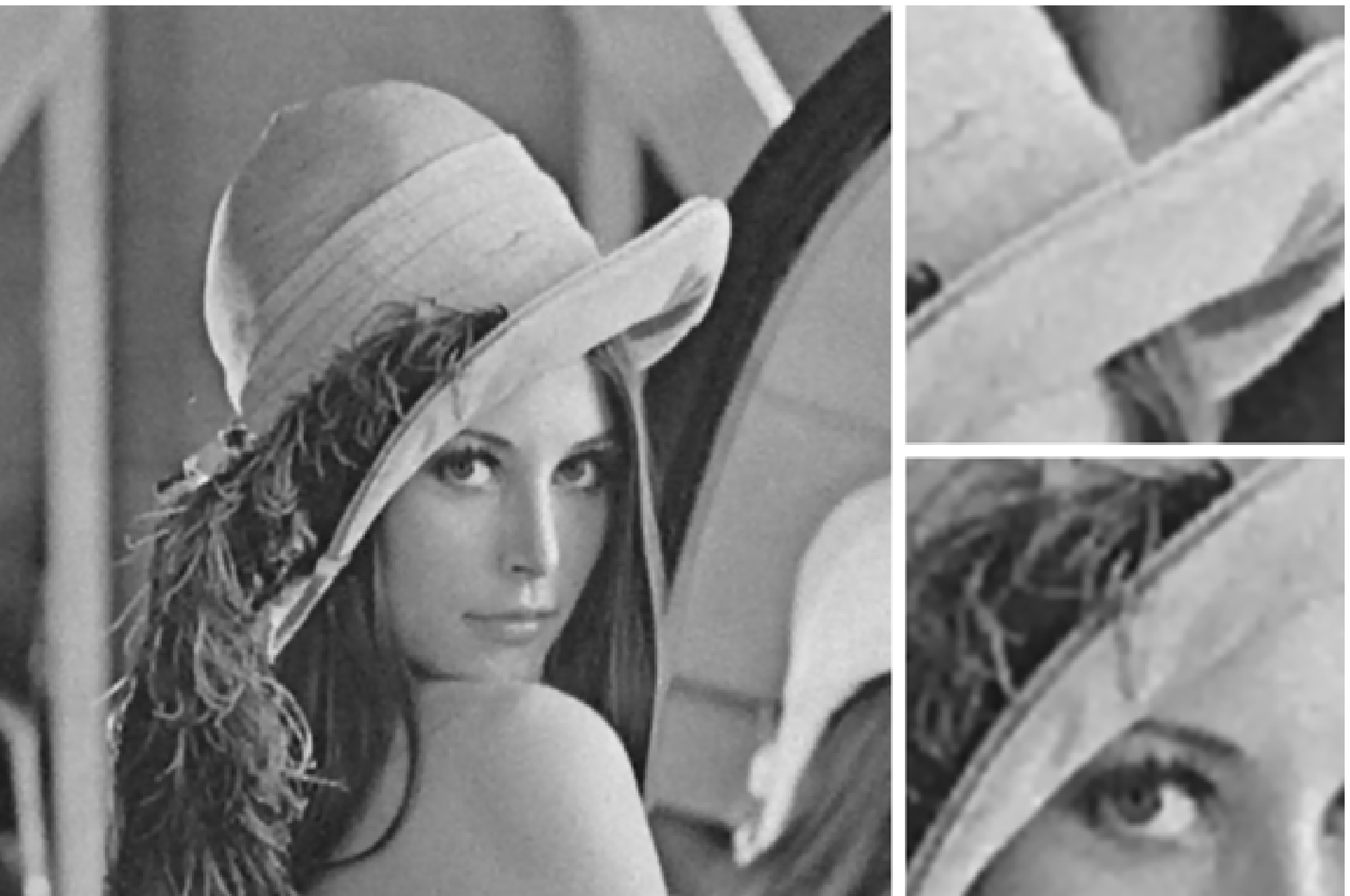} \hfill
\\ \vfill
    \hfill \includegraphics[width=0.49\linewidth]{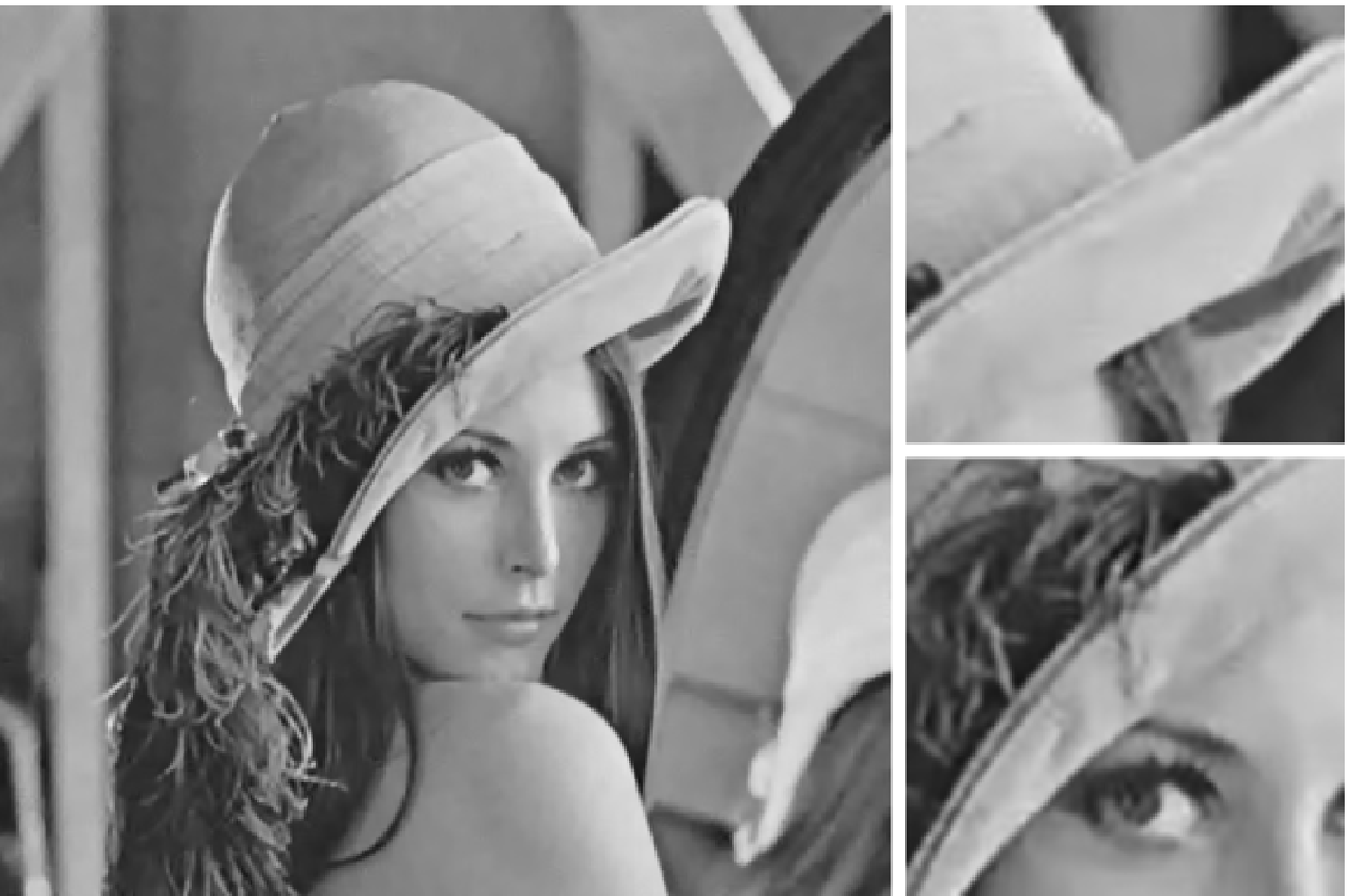} \hfill
    \includegraphics[width=0.49\linewidth]{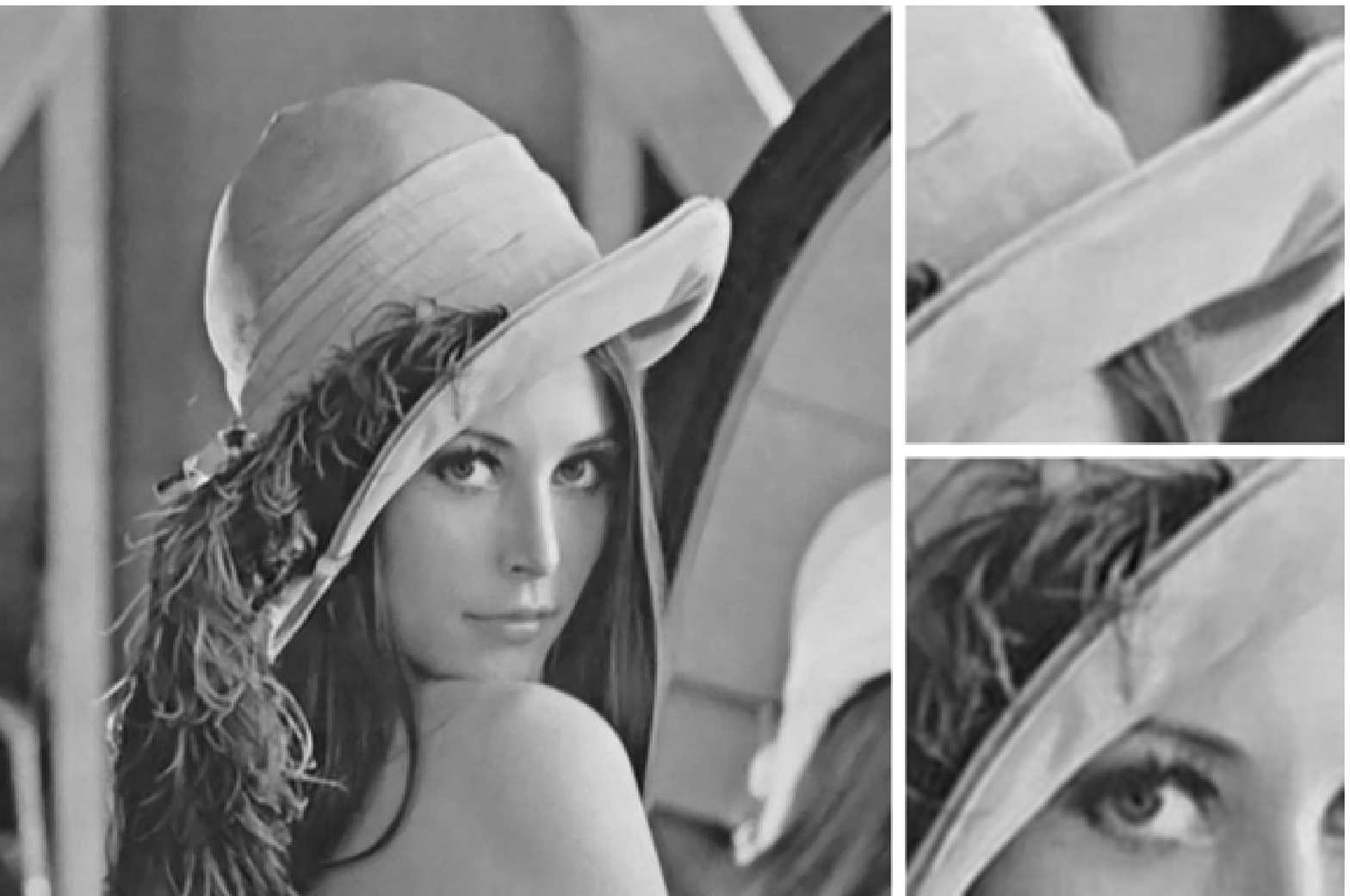} \hfill

  \caption{Examples of deblurring for the case 4. From top to bottom, left to right: original image, blurry image, Richardson-Lucy, sparse gradient \cite{sparseGradient}, SA-DCT \cite{foi2006}, our method.
 Best seen by zooming on a computer screen.}
  \label{fig:blur4}
\end{figure*}

For each blur level, we have generated pairs of blurry/sharp images from our
training database, and learned dictionaries of size $k=512$ elements. We have
observed that the results quality usually improves with the dictionary size,
$512$ being a good compromise between quality and computational cost.
Since our database is large, the parameters $\mu$ is always set to a negligible
value, $\mu=10^{-8}$. The size of patches $m_s$ and $m_b$ are respectively
set to $7$ and $11$ for all experiments.
The only parameter which should be carefully tuned to obtain good results
is the regularization parameter $\lambda$.
Following \cite{BM3D,foi2006,SVGSM}, we have manually chosen a value of~$\lambda$ via a rough grid search for each type of blur and used it for every image.
We report quantitative results in Table \ref{table:resultsBlur} in terms of improvement in signal-to-noise ratio (ISNR),\footnote{
Denoting by \textrm{MSE} the mean-squared-error
for images whose intensities are between $0$ and $255$, the
\textrm{PSNR} is defined as $\textrm{PSNR}=10\log_{10}( 255^2 /
\textrm{MSE} )$ and is measured in dB. A gain of $1$dB reduces
the \textrm{MSE} by approximately $20\%$.} and compare our method
to the classical Richardson-Lucy algorithm~\cite{richardson},
and to recent state-of-the-art methods~\cite{BM3D,foi2006,levin}. A few values are missing in the table: these experiments were not done by the authors of the papers. 
We observe that our method is competitive or better than the state of the art in experiments $2,3,4,5,6$, 
where the supports of the blur kernels are relatively small. On the contrary,
our algorithm is significantly behind other approaches in the case 1,
probably because our patches are too small compared to the kernel size. The simple linear model while not at the state of the art, is giving surprisingly good results for most of the blurs. 
Its combination with the dictionaries shows a significant improvement, leading to state-of-the-art performances. Qualitative
results are presented in Figures~\ref{fig:blur2}, \ref{fig:blur3} and \ref{fig:blur4}.          
\begin{table*}
\label{table:resultsBlur} 
\centering
\caption{Isotropic deblurring results in ISNR (PSNR improvement). For each image/experiment, the best result is in bold. Four values are missing: the results for this experiment
were taken from \cite{yu2010}, who does not test on the exact same set of images than us.}
\setlength{\tabcolsep}{1.5pt}
 \begin{tabular}{|| l || c | c | c | c | c | c ||c |c|c|c|c|c||}                                        
             
\hline   
\hline
&Exp. 1 & Exp. 2 & Exp. 3 & Exp. 4 & Exp. 5 & Exp .6  & Exp. 1 & Exp. 2 & Exp. 3 & Exp. 4 & Exp. 5 & Exp .6 \\
\hline
\hline
& \multicolumn{6}{| c ||}{\textit{Cameraman}}  & \multicolumn{6}{| c ||}{\textit{Lena}} \\
\hline 
PSNR input image & 20.76 & 22.35 & 22.29 & 24.7 & 25.53 & 23.44 & 25.84 & 27.57 & 27.35 & 29.00 & 30.74 & 28.97\\
\hline
Richardson-Lucy \cite{richardson} & 4.47 & 5.53 & 3.58 & 0.49 & 1.21 & 1.04 & 4.80 & 5.29 & 2.71 & 0.02 & 0.26 & 0.53 \\
\hline 
Sparse gradient \cite{sparseGradient} & 7.73 & 6.89 & 4.78 & 2.24 & 2.64 & 2.70 & 7.02 & 2.83 & 5.44 & 4.06 & 3.30 & 3.33 \\
\hline 
SA-DCT \cite{foi2006} & \textbf{8.55} & 8.11 & 6.33 & 3.37 & - & - & 7.79 & 7.55 & 6.10 & 4.49 & 3.56 & 3.46 \\
\hline
BM3D \cite{BM3D} & 8.34 & 8.19 & 6.40 & 3.34 & 3.73 & \textbf{3.83} & \textbf{7.97} & \textbf{7.95} & \textbf{6.53} & 4.81 & 4.18 & 4.12 \\
\hline
Linear & 3.34 & 7.72 & 6.00 & 3.20 & 3.47 & 2.69 & 3.58 & 7.30 & 5.82 & 4.64 & 3.89 & 3.58 \\
\hline
Linear + Dictionary   & 4.76 & \textbf{8.35} & \textbf{6.47} & \textbf{3.57} & \textbf{3.94} & 3.35 & 4.83 & 7.79 & 6.13 & \textbf{5.16} & \textbf{4.34} &  \textbf{4.17} \\
\hline
\hline               
 & \multicolumn{6}{| c ||}{\textit{House}} & \multicolumn{6}{| c ||}{\textit{Barbara}} \\
 \hline     
 PSNR input image & 24.11& 26.28 & 26.10 & 28.51 & 30.16 & 28.18 & 22.49 & 23.49 & 23.35 & 24.28 & 25.02 & 23.46 \\
 \hline
 Richardson-Lucy \cite{richardson} & 6.46 & 5.86 & 3.68 &  0.04  & 0.25 & 0.59 & 2.26 & 2.70 & 1.13 & -0.06 & 0.12 & 0.02 \\
 \hline
Sparse gradient \cite{sparseGradient} & 10.16 & 8.03 & 6.43 & 4.09 & 3.47 & 3.92 & 2.88 & 6.87 & 1.51 & 0.57 & 0.66 & 1.11 \\
\hline 
SA-DCT \cite{foi2006} & 10.5 & 9.02 & 7.74 & 4.99 & 4.14 & 4.21 & 4.79 & 5.45 & 2.54 & 1.31 & - & -\\
\hline
 Dabov et al. \cite{BM3D} & \textbf{10.85} & \textbf{9.32} & \textbf{8.14} & 5.13 & 4.79 &  \textbf{5.30} & \textbf{5.86} & \textbf{7.80} & 3.94 & 1.90 & \textbf{3.17} & \textbf{1.94} \\
\hline
Linear & 4.25 & 8.90 & 7.58 & 5.22 & 4.51 & 4.26 & 2.39 & 7.18 & 4.27 & 1.86 & 2.89 & 1.56 \\
\hline
Linear + Dictionary & 6.99 & \textbf{9.32} & 7.71 & \textbf{5.74} & \textbf{4.98} & 5.09 & 2.65 & 7.64 & \textbf{4.59} & \textbf{2.00} & 3.11 & 1.70 \\
 \hline
 \hline                     
 \end{tabular}
 \end{table*}
  
\subsection{Astronomical Images}
\label{sec:astronomy}
Our method is not designed specifically for the restoration of natural images. It adapts itself to the training set and can so be applied on various data. 
This versatility is illustrated here on astronomical imaging, which is a field where non-blind deblurring has had a major industrial impact. The experiment
setting is based on a classical astronomical case. A star image has to be recovered from a blurred and noisy version of it. The blur kernel is the Hubble Space Telescope kernel as given 
in~\cite{starck}. The additive noise is Gaussian. The training set is constructed from several others star images. 
 
\begin{figure*}
    \includegraphics[width=0.32\linewidth]{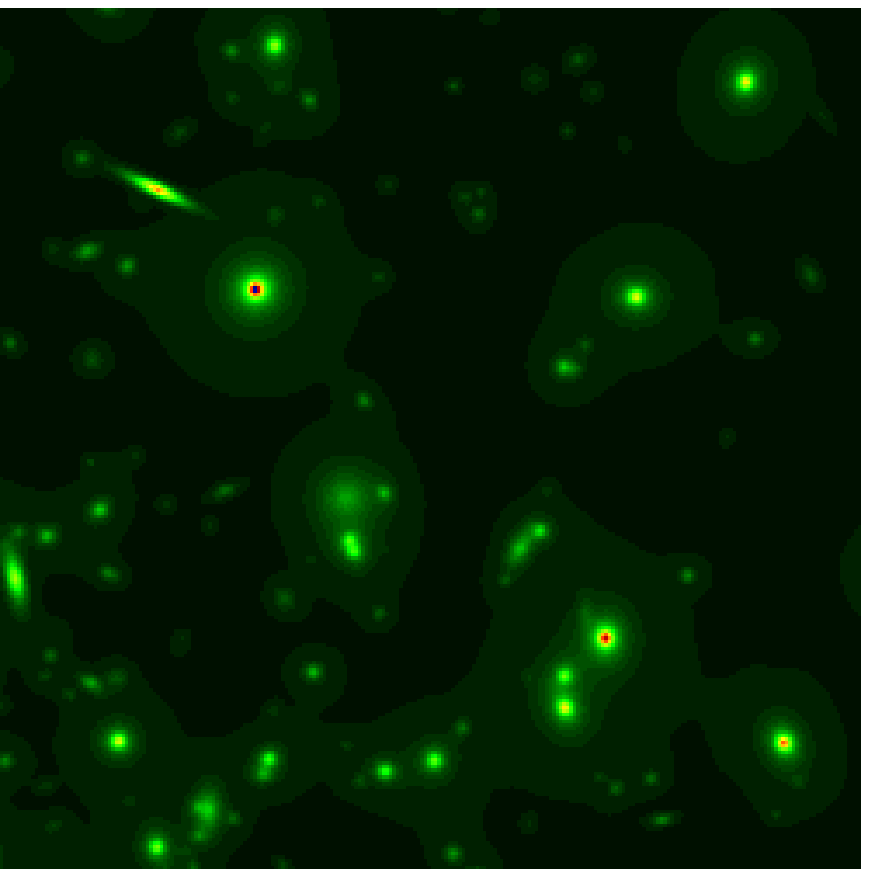} \hfill
    \includegraphics[width=0.32\linewidth]{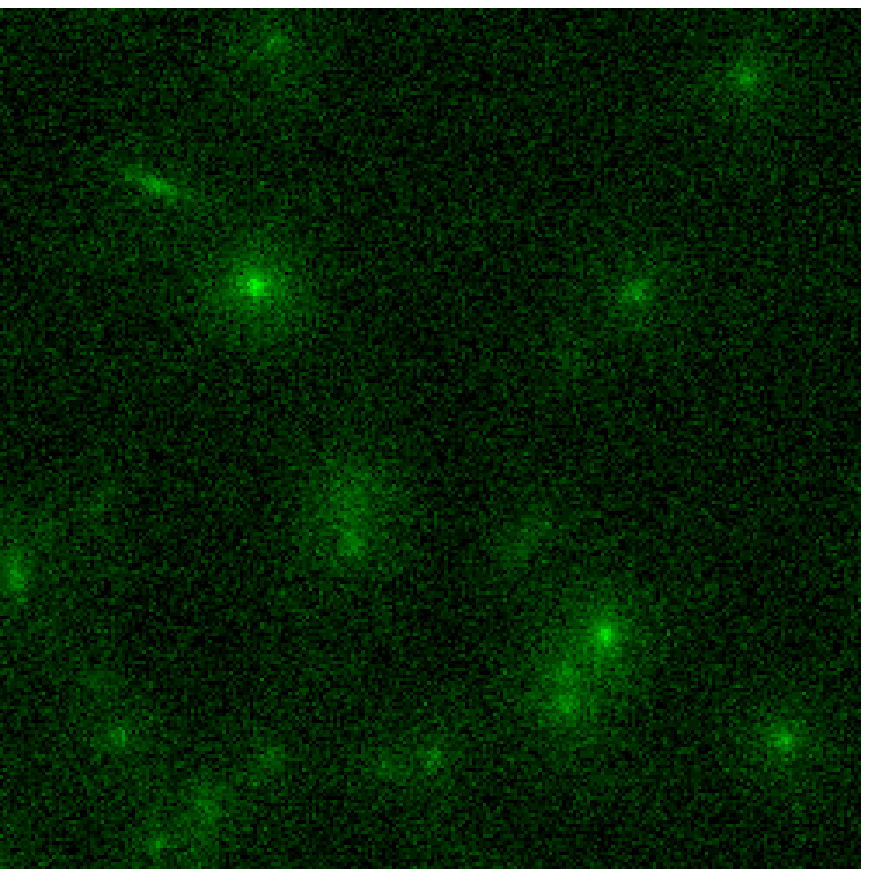} \hfill
     \includegraphics[width=0.32\linewidth]{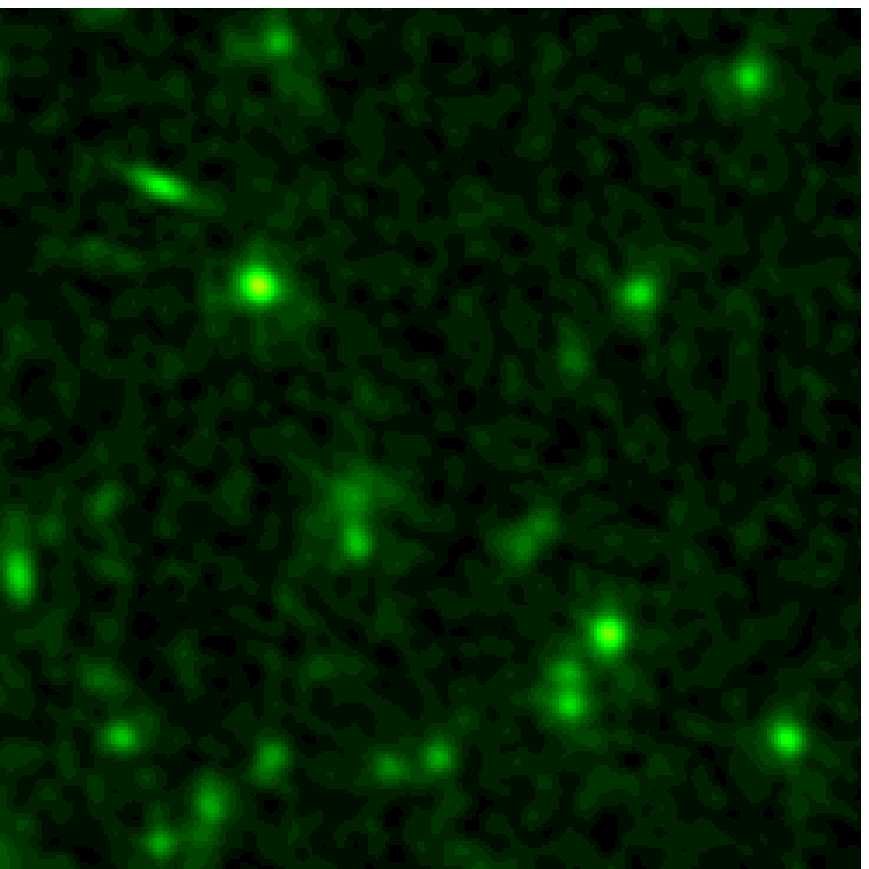} \\ \vfill
     \includegraphics[width=0.32\linewidth]{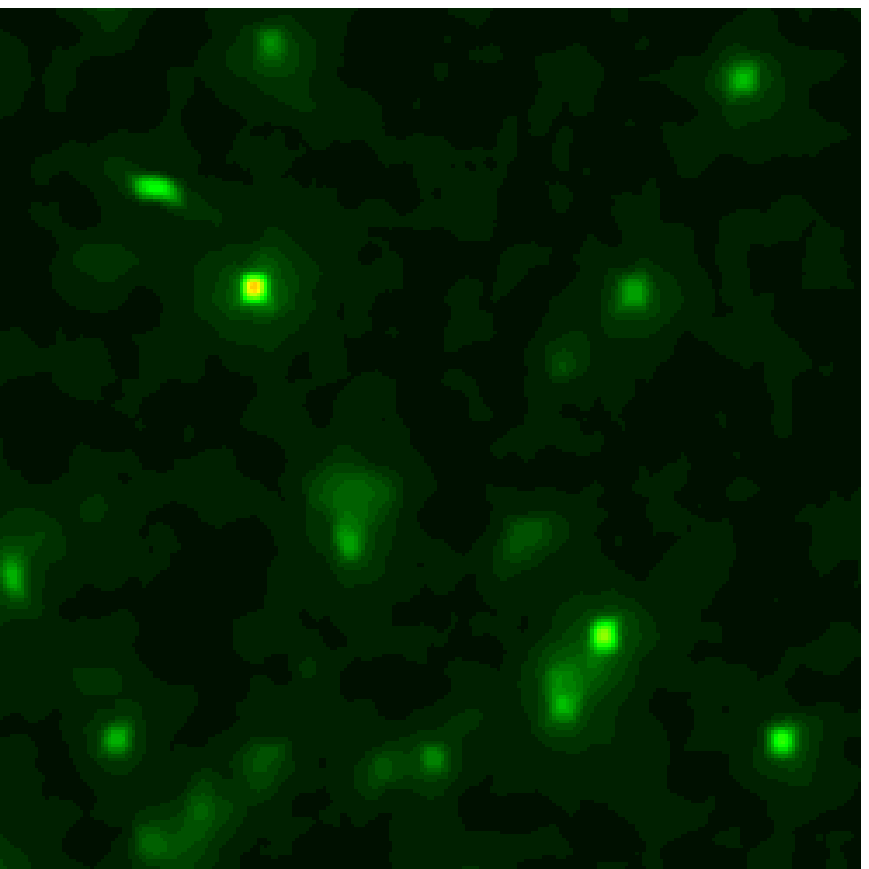} \hfill
     \includegraphics[width=0.32\linewidth]{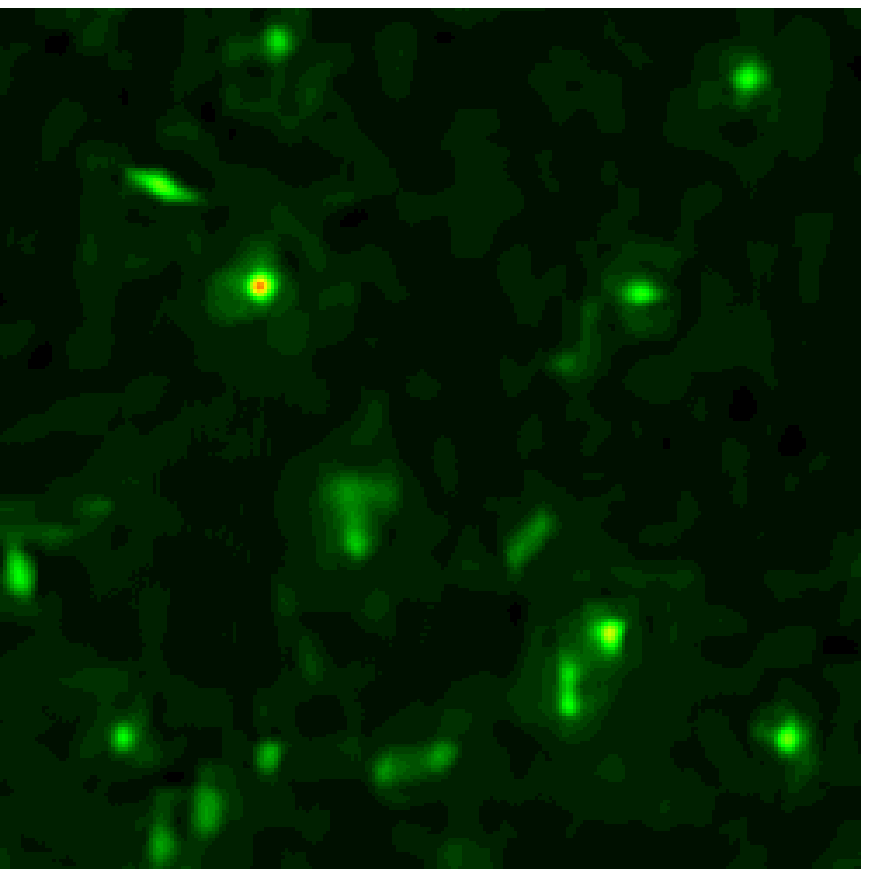} \hfill
    \includegraphics[width=0.32\linewidth]{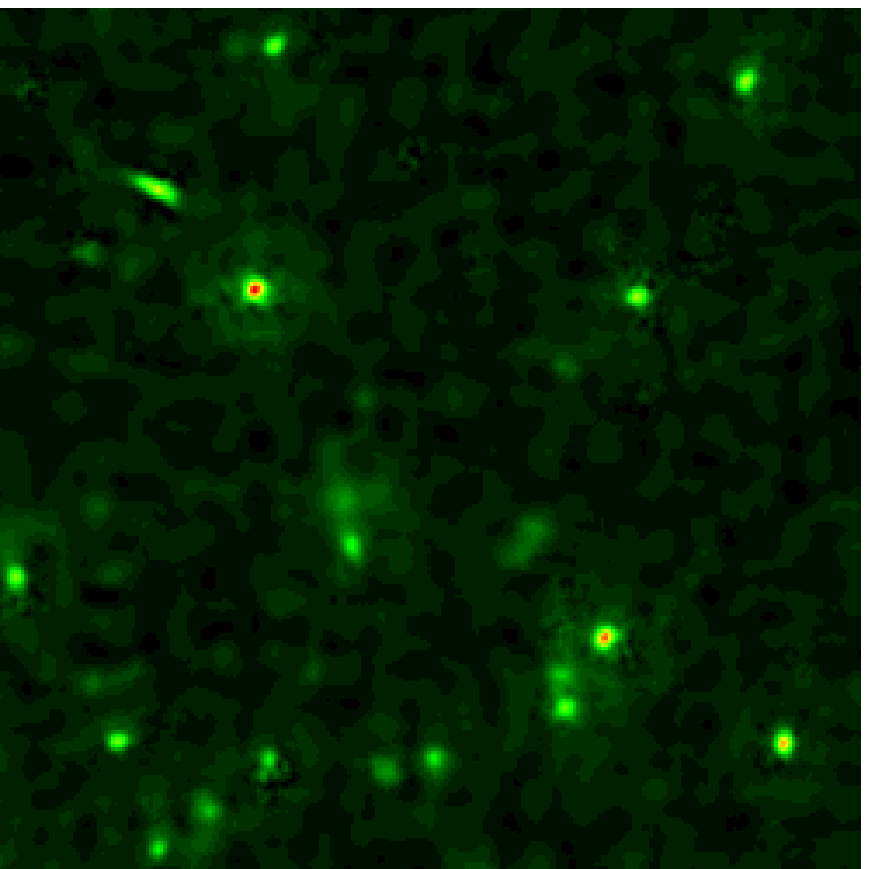}
    \caption{Example of deblurring of an astronomical image. Between parenthesis is indicated the PSNR. First line: Original, 
blurred and noisy image (25.3) , Wiener deblurring (29.6). Second line: sparse gradient \cite{sparseGradient} (31.3), BM3D~\cite{BM3D} (30.8), our method (33.5). Best seen in color.}
    \label{fig:astronomy}
 \end{figure*}

 Figure~\ref{fig:astronomy} presents the results with several deblurring algorithm. Our method result is quantitatively better than the other algorithms: While
 the two algorithms adapted to natural images~\cite{BM3D,sparseGradient} gives a PSNR of 30.8 and 31.3, our method gets 33.5. 
 In particular, our algorithm manages to recover really high values on the brightest stars. This is not surprising,
several of these algorithms use priors that do not fit well astronomical images, but it validates the capability of our method to adapt to various data.

\subsection{Non-Blind Deblurring with Anisotropic Kernels}
\begin{figure}
	\includegraphics[width=1\linewidth]{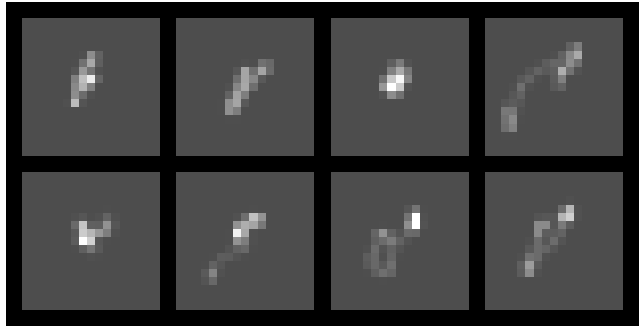}
	\caption{Anisotropic kernels from \cite{levin} used in our experiments.}
	\label{fig:anisoKernel}
\end{figure}

While deblurring isotropic blurs is sufficient in many applications, anisotropic blurring appears in practical cases, e.g., camera-shaking blur. To test our algorithm on this setting, we used 
the kernels from the database by Levin et al.~\cite{levin}. The local nature of our algorithm makes computationally challenging the treatment of large blurs and so we only worked with 
downsampled versions of the proposed kernels (by a factor 2). The 8 kernels used are shown in Figure~\ref{fig:anisoKernel}. White Gaussian noise of variance 2 is added to 
the blurry images before deblurring. We compare in Table~\ref{table:resultsBlurAnisotrop}  with the sparse-gradient-based algorithm from Levin et al.~\cite{sparseGradient} which is,
 to the best of our knowledge, the one giving the state-of-the-art results for this type of kernels. 

Our method does significantly worse than \cite{sparseGradient} on three of these kernels: there are the ones where the kernel is large and we think it is probably due to the locality of our predictor. 
For the $8$ kernels, we worked with patches of size $13$ and it might be not sufficient for too big kernels. 

\begin{table}
\begin{center}
\caption{Anisotropic deblurring results in mean ISNR (PSNR improvement) over 5 images. The kernels used are downsampled versions of those from~\cite{levin}.}
\label{table:resultsBlurAnisotrop}
\begin{tabular}{|c || c | c | c | c |}
\hline
\hline
Kernel & 1 & 2 &3 &4  \\
\hline
Sparse gradient~\cite{sparseGradient} & 9.04 & 6.91 & 7.49 & \textbf{10.67}  \\
\hline
Ours & \textbf{10.67} & \textbf{7.17}  &  \textbf{9.02} & 6.63  \\
\hline
\hline
Kernel &5 &6 &7 &8 \\
\hline
Sparse gradient~\cite{sparseGradient} & 8.64 & 9.18  & \textbf{11.15}  & \textbf{10.24} \\
\hline
Ours &  \textbf{10.52} & \textbf{10.03} &  9.64 &  7.75 \\
\hline
\hline

\end{tabular}

\end{center}	
\end{table}

\subsection{Digital Zoom}
Following the same experimental protocol than for the deblurring experiments, we have
evaluated our method for the digital zooming task. The dictionary size 
is $k=512$, and the patch sizes are $m_b=11$ and $m_s=7$.
Digital zooming is usually done on good quality images, with a very small noise: for this reason we use a small
regularization parameter $\lambda$, which is set to $0.005$.

It is always difficult to evaluate quantitatively the results of digital zoom algorithms.
Indeed, upsampling and downsampling methods are often subject to sub-pixel misalignments, which 
are visually imperceptible, but make important mean square error differences. Moreover, the antialiasing filter that has to be applied during the downsampling is rarely detailed, 
making comparisons difficult. For this experiment, we used the Matlab function \textit{imresize} with a bicubic interpolation to create the low-resolution images. 
The choice of the antialiasing, which allows to create the training set, is really important. With a too strong antialiasing our method might sharpen too much the images, while with a weak
antialiasing it might not deblur enough.

We compare quantitatively with the method from Yang et al. \cite{YiMa} that also uses dictionaries, proving the efficiency of the 
discriminative approach. The dictionaries sizes are the same as ours (512), and the parameter $\lambda$ is chosen on a validation set of images. 
This method works in two steps, first, it predicts a high-resolution image from a filtered version of the low resolution one 
using pairs of dictionaries, then, the image is cleaned using a backprojection. We compare the results at both steps with our method in Table \ref{tab:super}.

\begin{figure*}
  \includegraphics[width=0.24\linewidth]{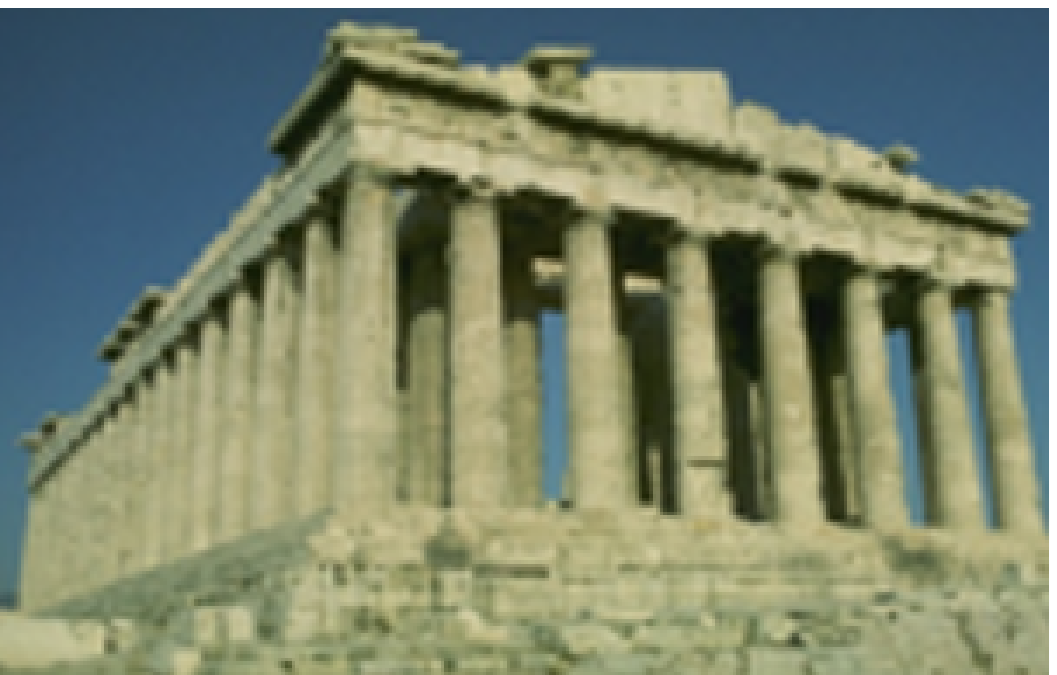} \hfill
   \includegraphics[width=0.24\linewidth]{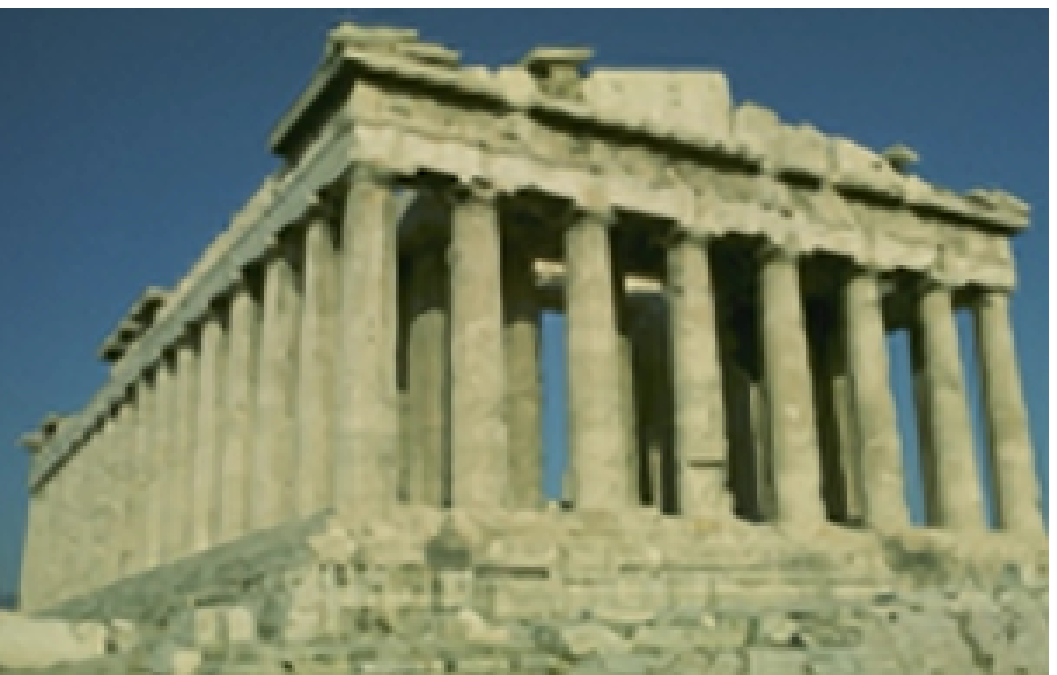} \hfill
   \includegraphics[width=0.24\linewidth]{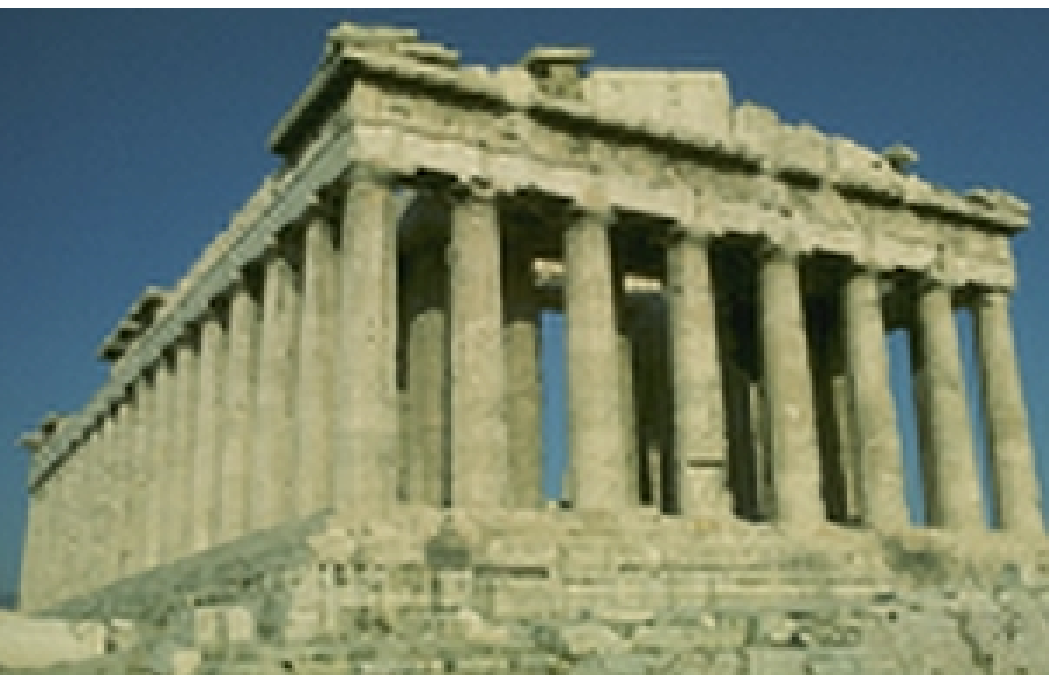} \hfill
   \includegraphics[width=0.24\linewidth]{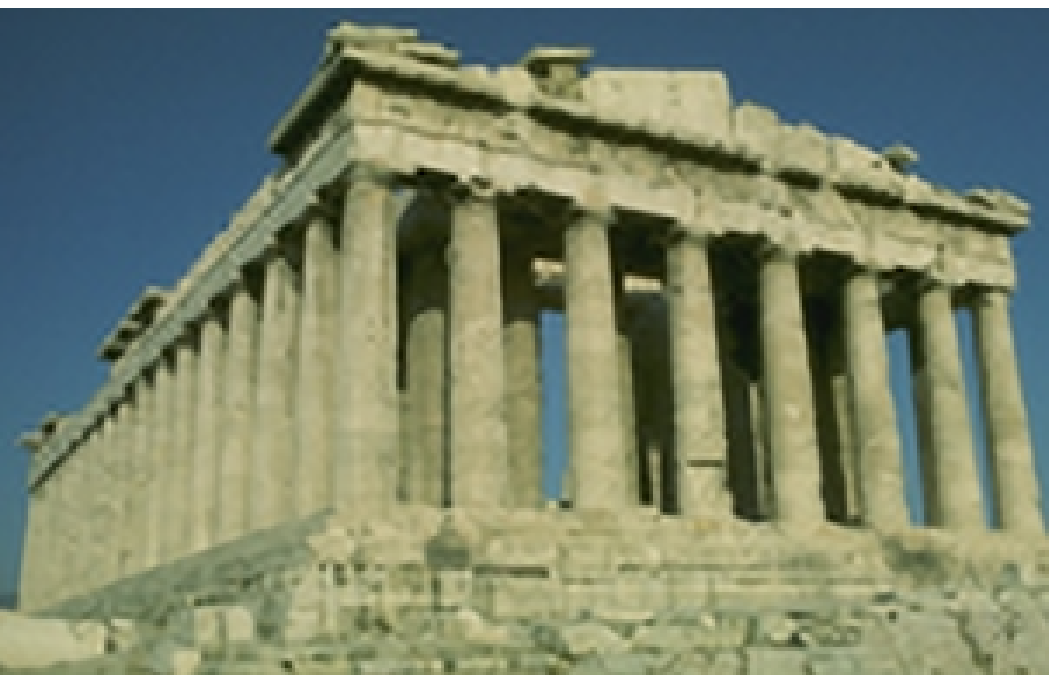}
\\ \vfill
   \includegraphics[width=0.24\linewidth]{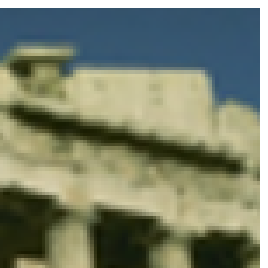} \hfill
   \includegraphics[width=0.24\linewidth]{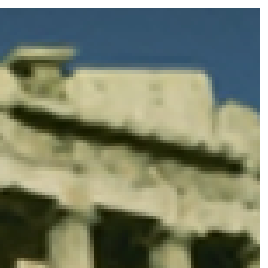} \hfill
   \includegraphics[width=0.24\linewidth]{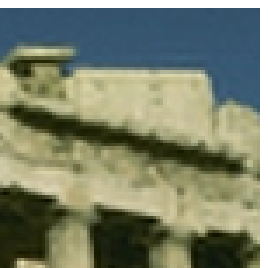} \hfill
   \includegraphics[width=0.24\linewidth]{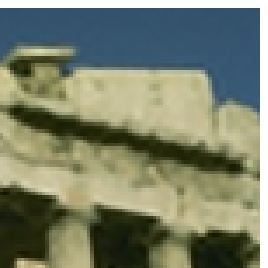} 
   \caption{Digital zoom by a factor 2. The top line shows the full image and the bottom one a zoom on one section. From left to right: bicubic interpolation, 
Yang et al.~\cite{YiMa} (dictionary only), Yang et al.~\cite{YiMa} with backprojection, our results}
   \label{fig:super1}
\end{figure*}

Our method outperforms the full method from Yang et al.~\cite{YiMa} by a small margin. But their results obtained only with dictionaries are significatively worse than
ours. The discriminative learning of the dictionaries and the addition of the linear predictor improve greatly the results.
\begin{table}
\centering
\caption{Digital zoom (by a factor 2) quantitative results in PSNR. We present two values for Yang et al. method: the first one
is the result given by their dictionaries, the second one is obtained by adding a backprojection algorithm to the dictionaries. For each image, the best result is in bold.}
\setlength{\tabcolsep}{1.5pt}
 \begin{tabular}{|c || c | c  | c |}
\hline
\hline
  & Cubic spline & Yang et al. \cite{YiMa} & Ours \\
\hline
\hline
Lena & 31.91 & 32.13 / 33.06 & \textbf{33.31} \\
\hline
Girl & 31.44 & 31.48 / 31.93 & \textbf{32.00} \\
\hline
Flower & 38.48 & 38.69 / 39.59 & \textbf{39.92} \\
\hline
\hline
\end{tabular}
\label{tab:super}
\end{table}
Figure~\ref{fig:super1} compares our results with the ones of Yang et al. using
one image from~\cite{YiMa}. We have observed that both methods improve significantly
upon the bicubic interpolation and gives similar results (with the backprojection step 
for Yang et al.~\cite{YiMa} method).

\begin{figure*}
    \includegraphics[width=0.24\linewidth]{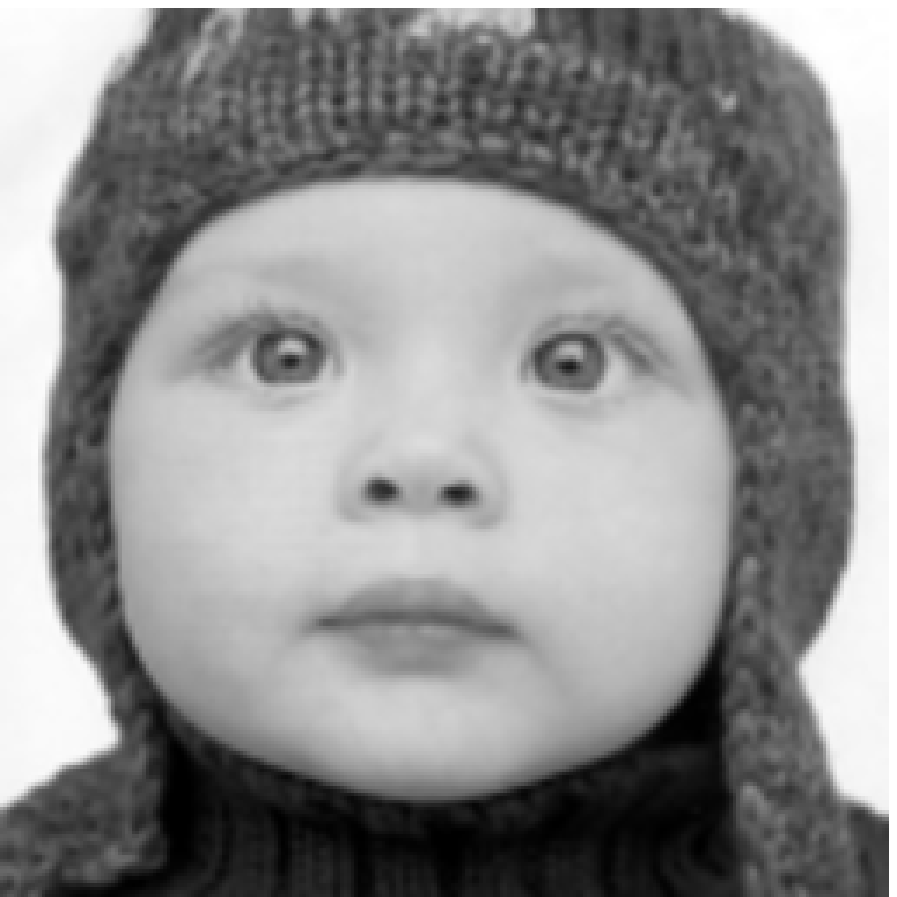} \hfill
    \includegraphics[width=0.24\linewidth]{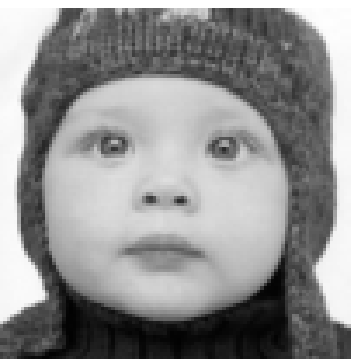} \hfill 
    \includegraphics[width=0.24\linewidth]{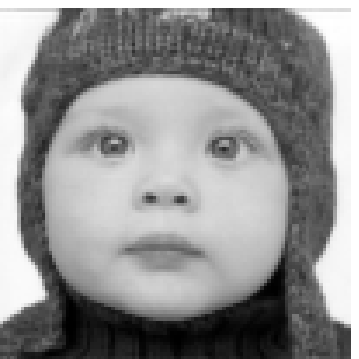} \hfill
    \includegraphics[width=0.24\linewidth]{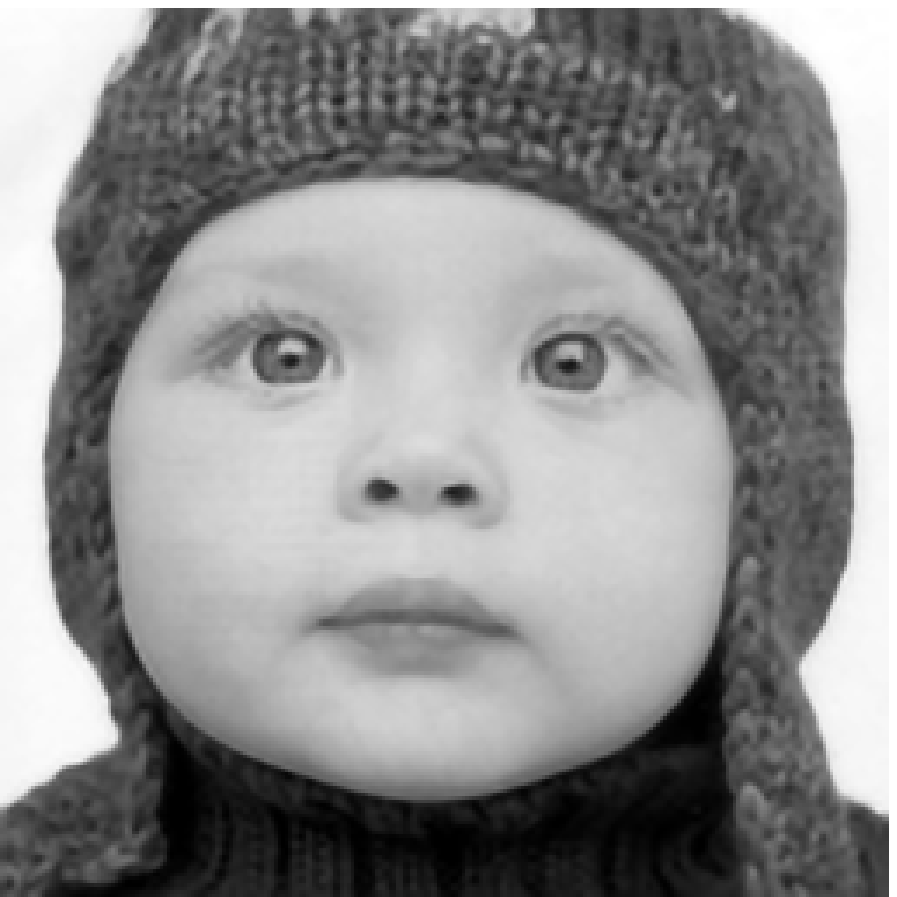}
\\ \vfill  
    \includegraphics[width=0.24\linewidth]{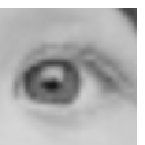} \hfill
    \includegraphics[width=0.24\linewidth]{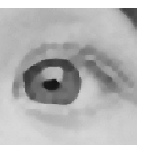} \hfill
    \includegraphics[width=0.24\linewidth]{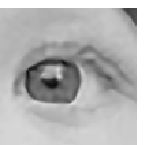} \hfill
    \includegraphics[width=0.24\linewidth]{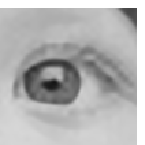} 
\\ \vfill
    \includegraphics[width=0.24\linewidth]{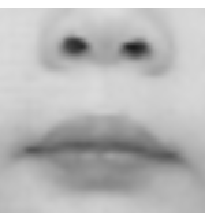} \hfill
    \includegraphics[width=0.24\linewidth]{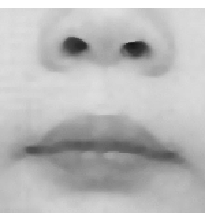} \hfill   
    \includegraphics[width=0.24\linewidth]{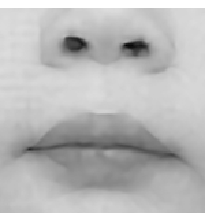} \hfill
    \includegraphics[width=0.24\linewidth]{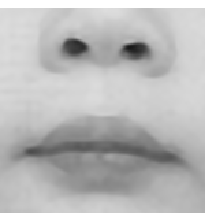} 
 
\caption{Digital zoom by a factor 4. From left to right: bicubic interpolation, Fattal et al~\cite{fattal}, Glasner et al.~\cite{glasdner}, our results. Best seen by zooming on a computer screen.}
   \label{fig:super2}
\end{figure*}

We have also compared qualitatively our method with others works:
In Figure~\ref{fig:super2}, we present digital zooming results (by a factor 4) obtained on one image from
\cite{fattal,glasdner}.  Our results are in general slightly better visually
than \cite{fattal} (see the texture of the baby's hat for instance), but
slightly behind~\cite{glasdner} in terms of sharpness of edges (e.g. the baby's mouth). On the other hand, Glasdner et al.~\cite{glasdner}'s algorithm 
reconstructs sometimes structures not present in the original image (e.g.,
square edges in the baby eye). In textured areas, we perform as good as~\cite{glasdner}.

\section{Conclusion} \label{sec:ccl}
In this paper, we have presented a new formulation for image deblurring and
digital zooming using a supervised formulation of dictionary learning combined with a linear predictor.
 With a stochastic gradient descent algorithm, our approach is efficient and
allows the use of millions of training samples.  Experiments on natural images
show that our method is competitive with the state of the art
for the non-blind deblurring and digital zooming tasks. Future work will
consist of extending the approach to the blind deblurring problem, where a blur
kernel has to be learned at the same time as the learned dictionaries, and
exploiting
self-similarities in images, which have proven to be very successful
for digital zooming~\cite{glasdner} and image denoising\cite{mairalNonLocal}.

\begin{acknowledgements}
This research was partially supported by the Agence Nationale de la Recherche (MGA Project) 
and the European Research Council (SIERRA and VideoWorld projects). In addition, Julien 
Mairal has been supported in part by the NSF grant SES-0835531 and NSF award CCF-0939370.
The authors would like to thanks Jean-Luc Starck for sharing the astronomical data used in subsection~\ref{sec:astronomy} and Jianchao Yang for providing us with his code of digital zooming. 
\end{acknowledgements}

\bibliographystyle{spmpsci}
\bibliography{manuscript}

\end{document}